\def\year{2022}\relax
\documentclass[letterpaper]{article} 
\usepackage{aaai22}  
\usepackage{times}  
\usepackage{helvet}  
\usepackage{courier}  
\usepackage[hyphens]{url}  
\usepackage{graphicx} 
\usepackage{natbib}  
\usepackage{caption} 
\DeclareCaptionStyle{ruled}{labelfont=normalfont,labelsep=colon,strut=off} 
\usepackage{algorithm}
\usepackage{algorithmic}

\usepackage{newfloat}
\usepackage{listings}
\floatstyle{ruled}
\newfloat{listing}{tb}{lst}{}
\floatname{listing}{Listing}

\usepackage{lipsum,booktabs}
\usepackage{subcaption}
\usepackage{multirow}
\usepackage{tabularx}
\usepackage{arydshln}
\usepackage{pifont}
\usepackage{algorithm}
\usepackage{algorithmic}
\usepackage{setspace}
\usepackage{lipsum}
\usepackage{graphics}
\usepackage{import}
\usepackage{booktabs} 
\usepackage{amsmath}
\usepackage{soul}
\usepackage{hyperref}       
\usepackage{url}            
\newcommand{\patchup}{\textit{PatchUp}}
\newcommand{\hardpatchup}{\textit{Hard PatchUp}}
\newcommand{\softpatchup}{\textit{Soft PatchUp}}
\newcommand{\cutmix}{\text{CutMix}}

\usepackage{amsmath,amssymb}
\DeclareMathOperator{\expectation}{\mathbb{E}}

\usepackage{graphicx}
\title{PatchUp: A Feature-Space Block-Level Regularization Technique for Convolutional Neural Networks}
\author{
    Mojtaba Faramarzi\textsuperscript{\rm 1 \rm 2},
    Mohammad Amini\textsuperscript{\rm 1 \rm 3},
    Akilesh Badrinaaraayanan,\textsuperscript{\rm 1 \rm 2},
    Vikas Verma\textsuperscript{\rm 1 \rm 2 \rm 4},
    Sarath Chandar \textsuperscript{\rm 1 \rm 5 \rm 6}
}
\affiliations{
    
    \textsuperscript{\rm 1}Mila - Quebec AI Institute,
    \textsuperscript{\rm 2}University of Montreal,
    \textsuperscript{\rm 3}McGill University,
    \textsuperscript{\rm 4}Aalto Univeristy, Finland,\\
    \textsuperscript{\rm 5}\'Ecole Polytechnique de Montréal,
    \textsuperscript{\rm 5}Canada CIFAR AI Chair

}

\usepackage{bibentry}

\makeatletter
\newcommand{\setlabel}[1]{\edef\@currentlabel{#1}\label}
\makeatother

\begin{document}

\maketitle

Large capacity deep learning models are often prone to a high generalization gap when trained with a limited amount of labeled training data. A recent class of methods to address this problem uses various ways to construct a new training sample by mixing a pair (or more) of training samples. We propose PatchUp, a hidden state block-level regularization technique for Convolutional Neural Networks (CNNs), that is applied on selected contiguous blocks of feature maps from a random pair of samples. Our approach improves the robustness of CNN models against the manifold intrusion problem that may occur in other state-of-the-art mixing approaches. Moreover, since we are mixing the contiguous block of features in the hidden space, which has more dimensions than the input space, we obtain more diverse samples for training towards different dimensions. Our experiments on CIFAR10/100, SVHN, Tiny-ImageNet, and ImageNet using ResNet architectures including PreActResnet18/34, WRN-28-10, ResNet101/152 models show that PatchUp improves upon, or equals, the performance of current state-of-the-art regularizers for CNNs. We also show that PatchUp can provide a better generalization to deformed samples and is more robust against adversarial attacks.
\section{Introduction}
\label{introduction}
\noindent Deep Learning (DL), particularly deep Convolutional Neural Networks (CNNs) have achieved exceptional performance in many machine learning tasks, including object recognition~\cite{krizhevsky}, image classification~\cite{krizhevsky,NIPS2015_Ren,Kaiming}, speech recognition~\cite{hinton} and natural language understanding \cite{seq2seq, transformers}. However, in a very deep and wide network, the network has a tendency to memorize the samples, which yields poor generalization for data outside of the training data distribution~\cite{arpit2017closer, goodfellow2016deep}. To address this issue, noisy computation is often employed during the training, making the model more robust against invariant samples and thus improving the generalization of the model~\cite{information_dropout}. This idea is exploited in several state-of-the-art regularization techniques. 

Such noisy computation based regularization techniques can be categorized into data-dependent and data-independent techniques~\cite{DBLP:Hongyu}. Earlier work in this area has been more focused on the data-independent techniques such as Dropout~\cite{JMLR:v15:srivastava14a}, Variational Dropout~\cite{gal2016theoretically} and ZoneOut~\cite{krueger2016zoneout}, Information Dropout~\cite{information_dropout}, SpatialDropout~\cite{DBLP:spatial_dropout}, and DropBlock~\cite{dropblock}. Dropout performs well on fully connected layers~\cite{JMLR:v15:srivastava14a}. However, it is less effective on convolutional layers~\cite{jonathan}. One of the reasons for the lack of success of dropout on CNN layers is perhaps that the activation units in the convolutional layers are correlated, thus despite dropping some of the activation units, information can still flow through these layers. SpatialDropout~\cite{jonathan} addresses this issue by dropping the entire feature map from a convolutional layer. DropBlock~\cite{dropblock} further improves SpatialDropout by dropping random continuous feature blocks from feature maps instead of dropping the entire feature map in the convolutional layers.

\begin{figure*}[tbh]
\centering
	\includegraphics[width=.9\textwidth]{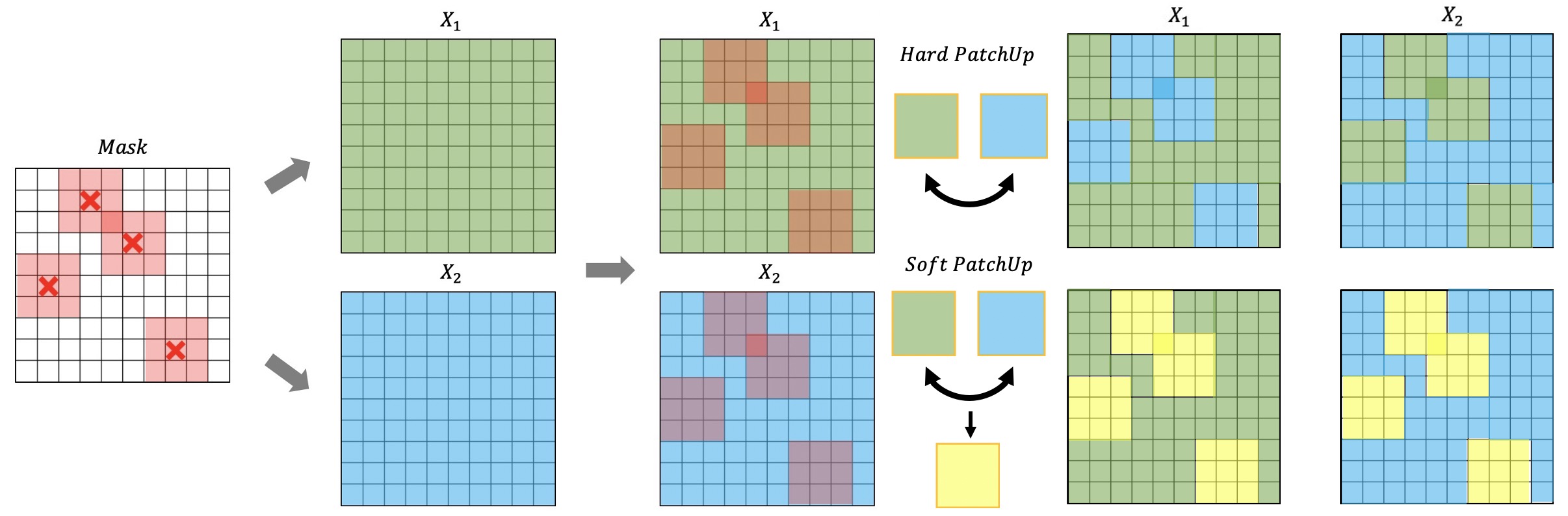} 
\caption{\patchup{} process for two hidden representations associated with two samples randomly selected in the mini-batch ($a,\;b)$. $X_{1} = g_{k}^{(i)}(a)$ and $X_{2} = g_{k}^{(i)}(b)$ where $i$ is the feature map index. Right top shows \hardpatchup{} output and the right bottom shows the interpolated samples with \softpatchup{}. The yellow continuous blocks represent the interpolated selected blocks.}
\label{fig:patchup_process}
\end{figure*}

Data-augmentation also is a data-dependent solution to improve the generalization of a model. Choosing the best augmentation policy is challenging. AutoAugment~\cite{cubuk2019autoaugment} finds the best augmentation policies using reinforcement learning with huge computation overhead. AugMix~\cite{hendrycks2020augmix} reduces this overhead by using stochasticity and diverse augmentations and adding a Jensen-Shannon Divergence consistency loss to training loss. Recent works show that data-dependent regularizers can achieve better generalization for CNN models. Mixup~\cite{mixup}, one such data-dependent regularizer, synthesizes additional training examples by interpolating random pairs of inputs $x_i$, $x_j$ and their corresponding labels $y_i$, $y_j$ as: 
\begin{equation}
\tilde{x}=\lambda x_{i}+(1-\lambda) x_{j} \text{\quad and \quad} \tilde{y}=\lambda y_{i}+(1-\lambda) y_{j},
\end{equation}
where $\lambda \in [0,1]$ is sampled from a Beta distribution such that $\lambda \sim Beta(\alpha, \alpha)$ and ($\tilde{x},\;\tilde{y}$) is the new example. By using these types of synthetic samples, Mixup encourages the model to behave linearly in-between the training samples. 
The mixing coefficient $\lambda$ in Mixup is sampled from a prior distribution. This may lead to the \textit{manifold intrusion problem}~\cite{DBLP:Hongyu}: the mixed synthetic example may \textit{collide} (i.e. have the same value in the input space) with other examples in the training data, essentially leading to two training samples which have the same inputs but different targets. To overcome the manifold intrusion problem, MetaMixUp~\cite{mai2019metamixup} used a meta-learning approach to learn  $\lambda$ with a lower possibility of causing such collisions. However, this meta-learning approach adds significant computation complexity. ManifoldMixup~\cite{manifold} attempts to avoid the manifold intrusion problem by interpolating the hidden states (instead of input states) of a randomly chosen layer at every training update. Recently, Puzzle Mix~\cite{kim2020puzzle} explicitly exploits an optimized masking strategy for the Input Mixup. It uses the saliency information and the underlying statistics of pair of images to avoid manifold intrusion problems at each batch training step. Puzzle Mix adds computation overhead at training time to find an optimal mask policy while improving the performance of the model in comparison to Mixup and ManifoldMixup. 

Different from the interpolation-based regularizers discussed above, Cutout~\cite{cutout} drops the contiguous regions from the image in the input space. This kind of noise encourages the network to learn the full context of the images instead of overfitting to the small set of visual features. 
CutMix~\cite{cutmix} is another data-dependent regularization technique that cuts and fills rectangular shape parts from two randomly selected pairs in a mini-batch instead of interpolating two selected pairs completely. Applying \cutmix{} at the input space improves the generalization of the CNN model by spreading the focus of the model across all places in the input instead of just a small region or a small set of intermediate activations. 
According to the \cutmix{} paper, applying \cutmix{} at the latent space, Feature \cutmix{}, is not as effective as applying \cutmix{} in the input space~\cite{cutmix}.

In this work, we introduce a feature-space block-level data-dependent regularization that operates in the hidden space by masking out contiguous blocks of the feature map of a random pair of samples, and then either mixes (\softpatchup{}) or swaps (\hardpatchup{}) these selected contiguous blocks. Our regularization method does not incur significant computational overhead for CNNs during training. \patchup{} improves the generalization of ResNet architectures on image classification task (on CIFAR-10, CIFAR-100, SVHN, and Tiny-ImageNet), deformed images classification, and against adversarial attacks. It also helps a CNN model to produce a wider variety of features in the residual blocks compared to other state-of-the-art regularization methods for CNNs such as Mixup, Cutout, CutMix, ManifoldMixup, and Puzzle Mix.
\section{PatchUp}
\label{patchup}
\patchup{} is a hidden state block-level regularization technique that can be used after any convolutional layer in CNN models. Given a deep neural network $f(x)$ where $x$ is the input, let $g_k$ be the $k$-th convolutional layer. The network $f(x)$ can be represented as $f(x) = f_k(g_k(x))$ where $g_k$ is the mapping from the input data to the hidden representation at layer $k$ and $f_k$ is the mapping from the hidden representation at layer $k$ to the output \cite{manifold}. In every training step, \patchup{} applies block-level regularization at a randomly selected convolutional layer $k$ from a set of intermediate convolutional layers. appx.\ref{appendix:random_k_selction} gives a formal intuition for selecting $k$ randomly.
\subsection{Binary Mask Creation}
\label{binary_mask}
Once a convolutional layer $k$ is chosen, the next step is to create a binary mask $M$ (of the same size as the feature map in layer $k$) that will be used to \patchup{} a pair of examples in the space of $g_k(x)$. The mask creation process is similar to that of DropBlock~\cite{dropblock}. The idea is to select contiguous blocks of features from the feature map that will be either mixed or swapped with the same features in another example. To do so, we first select a set of features that can be altered (mixed or swapped). This is done by using the hyper-parameter $\gamma$ which decides the probability of altering a feature. When we alter a feature, we also alter a square block of features centered around that feature which is controlled by the side length of this square block, $block\_size$. Hence, the altering probabilities are readjusted using the following formula \cite{dropblock}:
\begin{equation}
    \label{equ:gamma}
    \gamma_{adj} = \frac{\gamma \times \textit{(feature map's area)}}{\textit{(block's area)} \times \textit{(valid region to build block)}},
\end{equation}
where the area of the feature map and block are the $feat\_size^2$ and $block\_size^2$, respectively, and the valid region to build the block is $(feat\_size - block\_size + 1)^2$. 
For each feature in the feature map, we sample from $Bernoulli(\gamma_{adj})$. If the result of this sampling for feature $f_{ij}$ is 0, then $M_{ij} = 1$. If the result of this sampling for $f_{ij}$ is 1, then the entire square region in the mask with the center $M_{ij}$ and the width and height of the square of $block\_size$ is set to 0. Note that these feature blocks to be altered can overlap which will result in more complex block structures than just squares. The block structures created are called patches. Fig.\ref{fig:patchup_process} illustrates an example mask used by \patchup{}. The mask $M$ has 1 for features outside the patches (which are not altered) and 0 for features inside the patches (which are altered). See Fig.~\ref{fig:block_selection} and~\ref{fig:mask_cmp} in Appendix for more details.
\subsection{PatchUp Operation}
Once the mask is created, we can use the mask to select patches from the feature maps and either swap these patches (\hardpatchup{}) or mix them (\softpatchup{}).

Consider two samples $x_i$ and $x_j$. The \hardpatchup{} operation at layer $k$ is defined as follows:
\begin{equation}
       \mathbf{\phi_{hard}}(g_k(x_i), g_k(x_j)) = \mathbf{M} \odot g_k(x_i) + (\mathbf{1}- \mathbf{M}) \odot g_k(x_j), 
\end{equation}
where $\odot$ is known as the element-wise multiplication operation and $\mathbf{M}$ is the binary mask described in section ~\ref{binary_mask}. 
To define \softpatchup{} operation, we first define the mixing operation for any two vectors $a$ and $b$ as follows:
\begin{equation}
\label{equ:mix_lam}
\text{Mix}_\lambda(a,\;b) = \lambda \cdot a + (1 - \lambda) \cdot b,
\end{equation}
where $\lambda \in[0,1]$ is the mixing coefficient. Thus, the \softpatchup{} operation at layer $k$ is defined as follows:
\begin{equation}
\begin{aligned}
        \mathbf{\phi_{soft}}(g_k(x_i), g_k(x_j)) = &\mathbf{M} \odot g_k(x_i) + \text{Mix}_\lambda[((\mathbf{1}- \mathbf{M}) \\
        &\odot g_k(x_i)),\;((\mathbf{1}- \mathbf{M}) \odot g_k(x_j))],
\end{aligned}
\end{equation}
where $\lambda$ in the range of $[0,\;1]$ is sampled from a Beta distribution such that $\lambda \sim Beta(\alpha, \alpha)$. $\alpha$ controls the shape of the Beta distribution. Hence, it controls the strength of interpolation \cite{mixup}. \patchup{} operations are illustrated in Fig.~\ref{fig:patchup_process} (see more details in Algorithm~\ref{alg:patchup} in Appendix). 

\subsection{Learning Objective}
After applying the \patchup{} operation, the CNN model continues the forward pass from layer $k$ to the last layer in the model. The output of the model is used for the learning objective, including the loss minimization process and updating the model parameters accordingly. 
Consider the example pairs $(x_i, y_i)$ and $(x_j, y_j)$. Let $\phi_k = \phi(g_k(x_i), g_k(x_j))$ be the output of \patchup{} after the $k$-th layer. Mathematically, the CNN with \patchup{} minimizes the following loss function:
\begin{equation}
\begin{aligned}
\label{eq:swap_loss}
    L(f) = & \expectation_{(x_i,\;y_i) \sim P}\,
    \expectation_{(x_j,\;y_j) \sim P}\,
    \expectation_{\lambda \sim \text{Beta}(\alpha,\;\alpha)}
    \expectation_{k \sim \mathcal{S}} \\ &~~~~\text{Mix}_{p_{u}} [ \ell(f_{k}(\phi_k),\;y_i) \; , \ell(f_{k}(\phi_k),\;Y)]\; \\
    &~~~~ + \ell(f_{k}(\phi_k),\; W(y_i,\;y_j)),
\end{aligned}
\end{equation}
where $p_{u}$ is the fraction of the unchanged features from feature maps in $g_{k}(x_i)$ and $\mathcal{S}$ is the set of layers where \patchup{} is applied randomly. $\phi$ is $\phi_{hard}$ for \hardpatchup{} and $\phi_{soft}$ for \softpatchup{}.

$Y$ is the target corresponding to the changed features. In the case of \hardpatchup{}, $Y = y_j$ and in the case of \softpatchup{}, $Y = \textit{Mix}_{\lambda}(y_i, y_j)$. $W(y_i, y_j)$ calculates the re-weighted target according to the interpolation policy for $y_i$ and $y_j$. $W$ for \hardpatchup{} and \softpatchup{} is defined as follows:
\begin{align}
& W_{hard}(y_i, y_j) = \text{Mix}_{p_{u}}(y_i, y_j) \\  
& W_{soft}(y_i, y_j) = \text{Mix}_{p_{u}}(y_i,\;\text{Mix}_{\lambda}(y_i, y_j)).
\label{eq:target_reweighted}
\end{align}

The \patchup{} loss function has two terms where the first term is inspired from the CutMix loss function and the second term is inspired from the MixUp loss function (more detail in Appendix~\ref{appendix:learning_objective_details}).

\subsection{PatchUp in Input Space}
\label{sec:patchup_in_input}
By setting $k = 0$, we can apply \patchup{} to only the input space. When we apply \patchup{} to the input space, only the \hardpatchup{} operation is used, this is due to the reason that, as shown in ~\cite{cutmix},  swapping in the input space provides better generalization compared to mixing. Furthermore, we select only one random rectangular patch in the input space (similar to CutMix) because the \patchup{} binary mask is potentially too strong for the input space, which has only three channels, compared to hidden layers in which each layer can have a larger number of channels (more detail in section~``\ref{sec:patchup_loss_k}'').
\section{Relation to Similar Methods}
\label{discussion}
\label{patchup_dis}
\begin{figure*}[t]
\centering
\begin{minipage}{.47\textwidth}
  \centering
  \includegraphics[width=.4\linewidth]{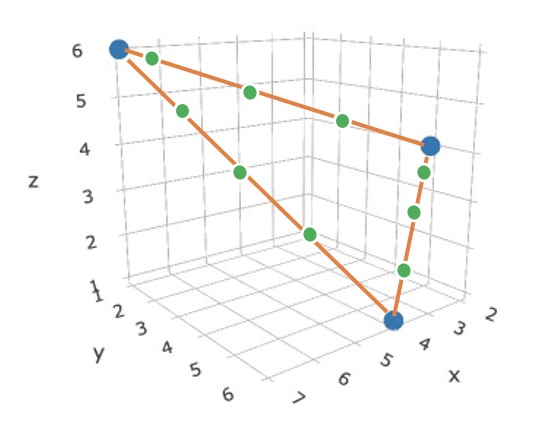} \ \
    \includegraphics[width=.4\linewidth]{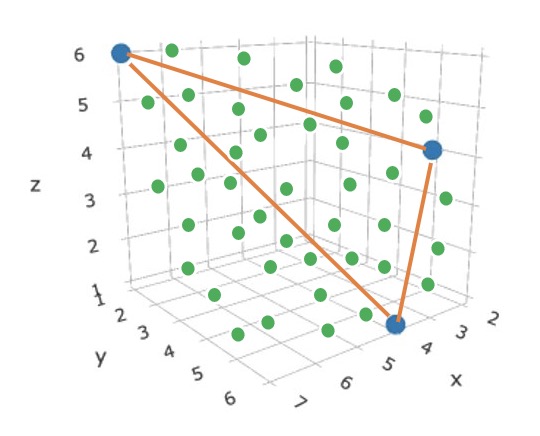}
    \caption{Left: ManifoldMixup interpolated samples for any combination of the three blue hidden states selected only from along orange line. Right: \patchup{} can produce interpolated hidden representations for these three hidden states in almost all possible places in all dimensions except the samples which lie directly on the orange lines.}
    \label{fig:diagram_mix_3d}
\end{minipage}%
\hfill
\begin{minipage}{.47\textwidth}
    \centering
\includegraphics[width=.46\linewidth]{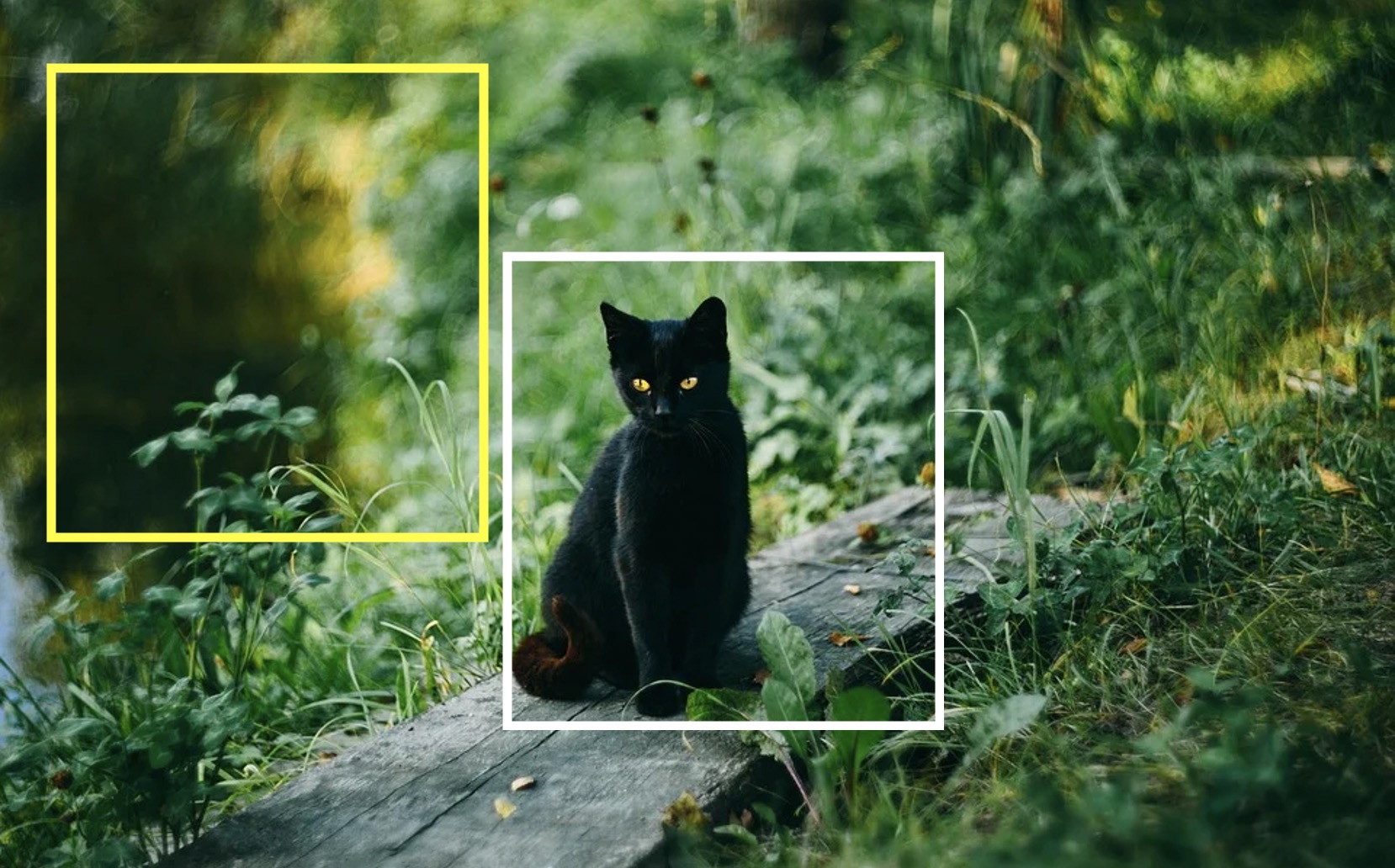} \hfill \ \
  \includegraphics[width=.46\linewidth]{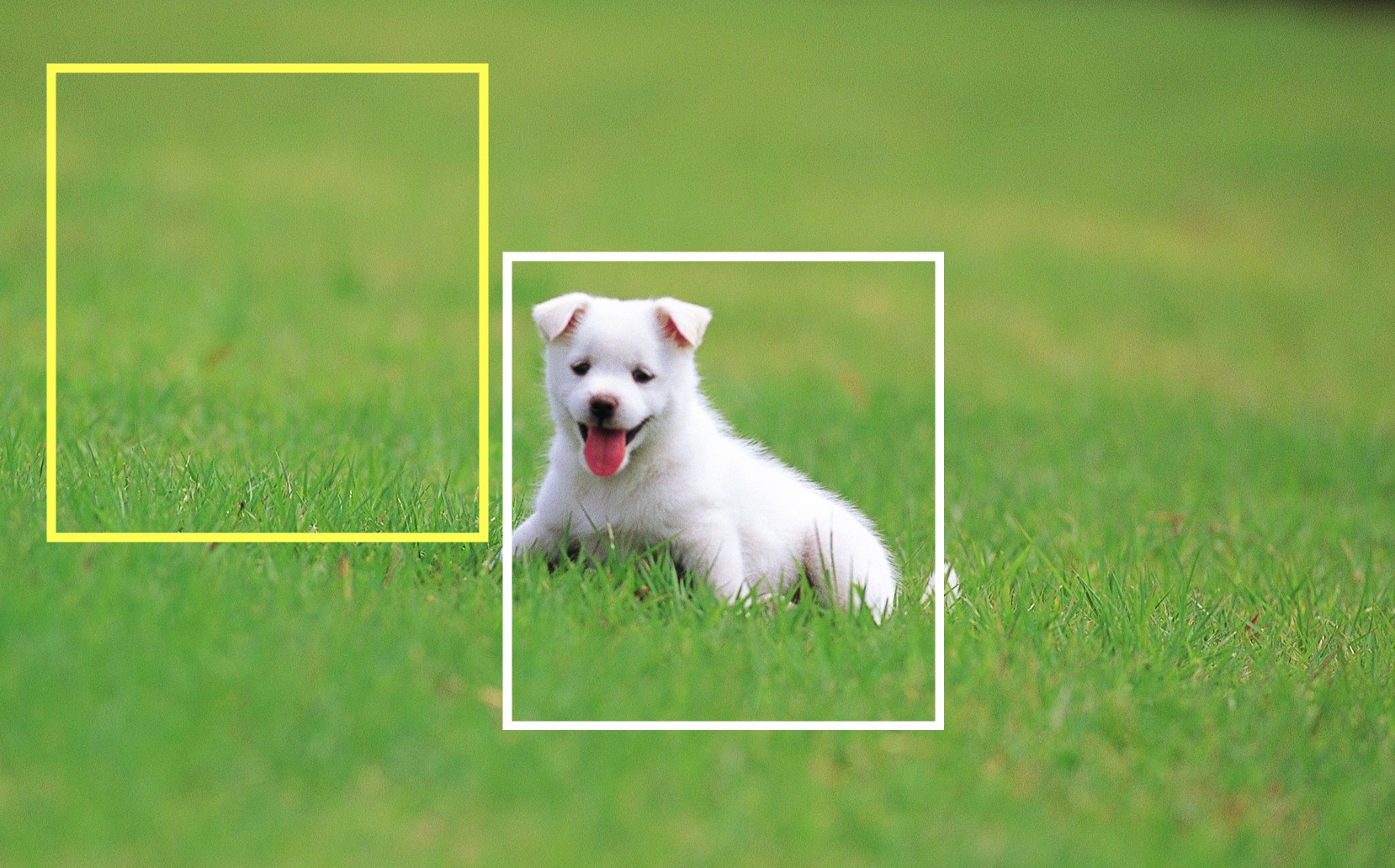}
    \caption{The two possible block selections from \cutmix{} for two samples (cat and dog) with a large background. Swapping a similar part of the background or an essential element correlated to the label in the selected images can have a negative effect on the \cutmix{} learning objective.}
    \label{fig:cat_dog}
\end{minipage}%
\end{figure*}

\paragraph{\patchup{} Vs. ManifoldMixup:}\label{para:patchup_vs_manifold} \patchup{} and ManifoldMixup improve the generalization of a model by combining the latent representations of a pair of examples. ManifoldMixup linearly mixes two hidden representations using Equation~\ref{equ:mix_lam}. \patchup{} uses a more complex approach ensuring that a more diverse subspace of the hidden space gets explored. To understand the behaviour and the limitation that exists in the ManifoldMixup, assume that we have a 3D hidden space representation as illustrated in Fig.\ref{fig:diagram_mix_3d}. It presents the possible combinations of hidden representations explored via ManifoldMixup and \patchup{}. Blue dots represent real hidden representation samples. ManifoldMixup can produce new samples that lie directly on the orange lines which connect the blue point pairs due to its linear interpolation strategy. But, \patchup{} can select various points in all dimensions, and can also select points extremely close to the orange lines. The proximity to the orange lines depends on the selected pairs and $\lambda$ sampled from the beta distribution. Fig.\ref{fig:diagram_mix_3d} is a simple diagrammatic description of how PatchUp constructs more diverse samples. Appx.~\ref{appendix:sim_comparison} provides a mathematical and real experimental justification.

\paragraph{\patchup{} Vs. \cutmix{}:}The \cutmix{} cuts and fills the rectangular parts of the randomly selected pairs instead of using interpolation for creating a new sample in the input space. Therefore, \cutmix{} has less potential for a manifold intrusion problem, however, \cutmix{} may still suffer from a manifold intrusion problem. Fig.\ref{fig:cat_dog} shows two samples with small portions that correspond to their labels. In this example, if only the parts within the yellow bounding boxes are swapped, then the label does not change. However, if the parts within the white bounding boxes are swapped, then the entire label is swapped. In both scenarios, \cutmix{} only learns the interpolated target based on the fractions of the swapped part. In contrast, these scenarios are less likely to occur in \patchup{} since it works in the hidden representation space most of the time. Another difference between \cutmix{} and \patchup{} is how the masks are created. \patchup{} can create arbitrarily shaped masks while \cutmix{} masks can only be rectangular. Fig.\ref{fig:mask_cmp} (appx.) shows an example of \cutmix{} and \patchup{} masks in input space and hidden representation space, respectively. \cutmix{} is more effective than Feature-CutMix that applies \cutmix{} in the latent space~\cite{cutmix}. The learning objective of \patchup{} and the binary mask selection are both different from that of Feature-CutMix.   
Experiments
\section{Experiments}
\label{experiments}
This section presents the results of applying \patchup{} to image classification tasks using various benchmark datasets such as CIFAR10, CIFfAR100~\cite{dataset:cifar}, SVHN (the standard version with 73257 training samples)~\cite{netzer2011reading}, Tiny-ImageNet~\cite{chrabaszcz2017downsampled}, and  with various benchmark architectures such as PreActResNet18/34~\cite{he2016identity}, and ResNet101, ResNet152, and WideResNet-28-10 (WRN-28-10)~\cite{zagoruyko2017wide}\footnote{The code is available: \url{https://github.com/chandar-lab/PatchUp}}. We used the same set of base hyper-parameters for all the models for a thorough and fair comparison. 
The details of the experimental setup and the hyper-parameter tuning are given in appendix-E. We set $\alpha$ to $2$ in \patchup{}. 
\patchup{} has $patchup\_prob$, $\gamma$ and $block\_size$ as additional hyper-parameters. $patchup\_prob$ is the probability that \patchup{} is performed for a given mini-batch. Based on our hyper-parameter tuning, \hardpatchup{} yields the best performance with $patchup\_prob$, $\gamma$, and $block\_size$ as $0.7$, $0.5$, and $7$, respectively.
\softpatchup{} achieves the best performance with $patchup\_prob$, $\gamma$, and $block\_size$ as $1.0$, $0.75$, and $7$, respectively.

\subsection{Generalization on Image Classification}
Table~\ref{tab:tbl-cifar-svhn} shows the comparison of the generalization performance of \patchup{} with six recently proposed mixing-based or feature-level methods on the CIFAR10/100, and SVHN datasets. Since Puzzle Mix clearly showed that both CutMix and Puzzle Mix perform better than AugMix~\cite{kim2020puzzle}, we excluded it from our experiments. 
Tables 12 and 11 in appendix-I show test errors and NLLs.
Our experiments show that \patchup{} leads to a lower test error for all the models on CIFAR, SVHN, and Tiny-ImageNet with a large margin. Specifically, \softpatchup{} outperforms other methods on Tiny-ImageNet dataset using ResNet101/152, and WRN-28-10 followed by \hardpatchup{}. As explained in appendix-C and shown in Fig. 10, both \textit{Soft} and \hardpatchup{} produce a wide variety of interpolated hidden representations towards different dimensions. However, \softpatchup{} behaves more conservatively which helps to outperform other methods by a large margin in the case of a limited number of training samples per class and having more targets.
\begin{table*}[tbp!]
\setlength{\tabcolsep}{6pt}
\centering
    \begin{subtable}{.9\linewidth}
        \resizebox{\linewidth}{!}{
\begin{tabular}{lllllllll}
\toprule
   & \multicolumn{2}{c}{\shortstack[l]{ResNet101}}                           &                      & \multicolumn{2}{c}{\shortstack[l]{ResNet152}}                           &                      & \multicolumn{2}{c}{\shortstack[l]{WideResnet-28-10}}                          \\ \cline{2-3} \cline{5-6} \cline{8-9}
  & \multicolumn{1}{c}{\shortstack[l]{Error}} & \multicolumn{1}{c}{\shortstack[l]{Loss}} & \multicolumn{1}{c}{} & \multicolumn{1}{c}{\shortstack[l]{Error}} & \multicolumn{1}{c}{\shortstack[l]{Loss}} & \multicolumn{1}{c}{} & \multicolumn{1}{c}{\shortstack[l]{Eror}} & \multicolumn{1}{c}{\shortstack[l]{Loss}} \\ \cline{2-3} \cline{5-6} \cline{8-9}
\shortstack[l]{No Mixup}                      & $46.18 \pm 0.49$              &  $2.37 \pm 0.03$ &                      &  $45.22 \pm 0.32$               &  $2.28 \pm 0.08$ &                      &  $36.76 \pm 0.25$               &  $1.89 \pm 0.02$  \\ 
\shortstack[l]{Input Mixup ($\alpha$ = 1)}  &  $44.93 \pm 0.34$               &  $2.25 \pm 0.05$  &                      &  $44.65 \pm 0.11$                &  $2.23 \pm 0.04$ &                      &  $36.50 \pm 0.29$               &  $1.88 \pm 0.02$ \\ 
\shortstack[l]{ManifoldMixup ($\alpha$= 2)} &  $44.54 \pm 0.14$               &  $2.25 \pm 0.01$ &                      &  $44.13 \pm 0.31$               &  $2.23 \pm 0.02$  &                      &  $35.96 \pm 0.64$               &  $1.83 \pm 0.06$ \\ 
\shortstack[l]{Cutout}                        &  $45.78 \pm 0.35$              &  $2.35 \pm 0.01$  &                      &  $45.23 \pm 0.19$               &  $2.34 \pm 0.02$  &                      &  $35.85 \pm 0.11$               &  $1.83 \pm 0.01$  \\ 
\shortstack[l]{DropBlock}                     &  $46.98 \pm 0.53$               &  $2.27 \pm 0.10$  &                      &  $45.82 \pm 0.22$               &  $2.17 \pm 0.01$ &                      &  $36.97 \pm 0.38$                &  $1.89 \pm 0.02$ \\ 
\shortstack[l]{CutMix}                        &  $42.04 \pm 0.38$              &  $2.11 \pm 0.01$ &                      &  $41.71 \pm 0.71$               &  $2.08 \pm 0.03$ &                      &  $35.81 \pm 0.18$               &  $1.75 \pm 0.08$  \\ 
\shortstack[l]{Puzzle Mix}                        &  $42.25 \pm 0.19$             &  $2.13 \pm 0.01$ &                      &  $42.13 \pm 0.31$    &  $2.13 \pm 0.03$ &                      &  $33.43 \pm 0.22$    &  $1.65 \pm 0.05$ \\
\shortstack[l]{\softpatchup{}}                &  {$\mathbf{38.68 \pm 0.34}$}    &  $1.87 \pm 0.01$ &                      &  {$\mathbf{38.11 \pm 0.29}$}     &  $1.84 \pm 0.02$ &                      &  {$\mathbf{29.81 \pm 0.24}$}  &  $1.46 \pm 0.04$ \\ 
\shortstack[l]{\hardpatchup{}}                &  {\ul{$40.55 \pm 0.15$}}       &  $1.94 \pm 0.01$ &                      & {\ul{$40.31 \pm 0.18$}}    &  $1.93 \pm 0.01$ &                      &  {\ul{$33.12 \pm 0.11$}}   &  $1.57 \pm 0.05$ \\ \bottomrule
\end{tabular}}
\end{subtable}
 \captionsetup[subtable]{position = bottom}
    \captionsetup[table]{position=bottom}
    \caption{Classification on Tiny-ImageNet. Best performance result is shown in bold, second best is underlined (five times).}
    \label{tab:tbl-tiny_imagenet}
\end{table*}

\hardpatchup{} provides the best performance in the CIFAR and \softpatchup{} achieves the second-best performance except on the CIFAR10 with WRN-28-10 where Puzzle Mix provides the second-best performance. In the SVHN, ManifoldMixup achieves the second-best performance in PreActResNet18 and 34 where \hardpatchup{} provide the lowest top-1 error. \softpatchup{} performs reasonably well and comparable to ManifoldMixup for PreActResNet34 on SVHN and leads to a lower test error followed by \hardpatchup{} for WRN-28-10 in the SVHN. We observe that the Mixup, ManifoldMixup, and Puzzle Mix are sensitive to the $\alpha$ when we have more training classes. It is notable that using the same $\alpha$, that is used in CIFAR or SVHN, leads to worst performance than No-Mixup in Tiny-ImageNet (Table~\ref{tab:tbl-tiny_imagenet}) where others are almost stable (more details in the Appendix). PatchUp and other methods reach the reported performance with WRN-28-10 model on Tiny-ImageNet after about 23 hours of training using one GPU (V100). However, Puzzle Mix needs 53 hours for training (more in  Table 8-appx.).
\begin{table}[htpb!]
\setlength{\tabcolsep}{1pt}
\centering
    \begin{subtable}{1.\linewidth}
        \resizebox{\linewidth}{!}{
        \begin{tabular}{lllll}
        \toprule
        \shortstack[l]{PreActResNet18}                                   & \shortstack[l]{\textbf{CIFAR-10}}   &     & \shortstack[l]{\textbf{CIFAR-100}} & \shortstack[l]{\textbf{SVHN}}\\
        \shortstack[l]{}                                   & \shortstack[l]{Test Error}   &     & \shortstack[l]{Test Error} & \shortstack[l]{Test Error}\\\midrule
        \shortstack[l]{No Mixup}                         &  $4.80 \pm 0.14$                    &     & $24.62 \pm 0.36$           &  $3.04 \pm 0.09$    \\
        \shortstack[l]{Input Mixup ($\alpha$ = 1)}     &  $3.63 \pm 0.20$                    &     & $22.33 \pm 0.32$           &  $2.93 \pm 0.22$     \\
        \shortstack[l]{ManifoldMixup ($\alpha$ = 1.5)}    &  $3.39 \pm 0.05$                 &     &   $21.40 \pm 0.38$       &  {\ul{$2.44 \pm 0.06$}}     \\
        \shortstack[l]{Cutout}                           &  $4.22 \pm 0.05$                    &     & $23.39 \pm 0.19$          &  $2.79 \pm 0.12$     \\
        \shortstack[l]{DropBlock}                        &  $5.04 \pm 0.15$                    &     & $25.02 \pm 0.26$          &  $2.96 \pm 0.11$     \\
        \shortstack[l]{CutMix}                           &  $3.52 \pm 0.90$                    &     & $22.18 \pm 0.18$           &  $3.04 \pm 0.05$     \\
        \shortstack[l]{Puzzle Mix}                       &   $3.16 \pm 0.11$                   &     &       $20.65 \pm 0.21$   &   $2.66 \pm 0.04$    \\
        \shortstack[l]{\softpatchup{}}                   &  {\ul{$2.95 \pm 0.12$}}          &     & {\ul{$19.95 \pm 0.18$}} &  $2.55 \pm 0.06$     \\
        \shortstack[l]{\hardpatchup{}}                   &  $\mathbf{2.92 \pm 0.13}$             &     & $\mathbf{19.12 \pm 0.17}$  &  {$\mathbf{2.29 \pm 0.08}$}     \\ \midrule 
        \shortstack[l]{PreActResNet34} & & & \\ \midrule
        \shortstack[l]{No Mixup}                        &  $4.64 \pm 0.10$                     &     & $23.34 \pm 0.27$            & $3.09 \pm 0.66$                \\
        \shortstack[l]{Input Mixup ($\alpha$ = 1)}    &  $3.26 \pm 0.08$                     &     & $21.00 \pm 0.44$            & $2.86 \pm 0.10$                \\
        \shortstack[l]{ManifoldMixup ($\alpha$ = 1.5)}  &  $2.93 \pm 0.06$                   &     & $18.72 \pm 0.31$            & {\ul{$2.42 \pm 0.43$}}      \\
        \shortstack[l]{Cutout}                          &  $3.69 \pm 0.14$                     &     & $22.42 \pm 0.08$            & $2.65 \pm 0.15$                \\
        \shortstack[l]{DropBlock}                       &  $4.95 \pm 0.19$                     &     & $23.74 \pm 0.13$            & $3.10 \pm 0.08$                \\
        \shortstack[l]{CutMix}                          &  $3.33 \pm 0.07$                     &     & $19.94 \pm 0.14$            & $2.66 \pm 0.05$                \\
        \shortstack[l]{Puzzle Mix}                       & $2.99 \pm 0.07$                     &     &        $19.97\pm 0.23$      & $2.45 \pm 0.08$              \\
        \shortstack[l]{\softpatchup{}}                  &  {\ul{$2.57 \pm 0.06$}}           &     & {\ul{$18.63 \pm 0.15$} } &  $2.47 \pm 0.08$               \\
        \shortstack[l]{\hardpatchup{}}                  &  {$\mathbf{2.53 \pm 0.05}$}          &     & {$\mathbf{17.69 \pm 0.13}$} & {$\mathbf{2.12 \pm 0.02}$}     \\ \midrule 
        \shortstack[l]{WideResNet-28-10} & & & \\ \midrule
        \shortstack[l]{No Mixup}                        &  $4.24 \pm 0.14$                     &     & $22.44 \pm 0.23$            & $2.83 \pm 0.08$     \\
        \shortstack[l]{Input Mixup ($\alpha$ = 1)}    &  $3.27 \pm 0.35$                     &     & $18.73 \pm 0.15$            & $2.64 \pm 0.16$     \\
        \shortstack[l]{ManifoldMixup ($\alpha$ = 1.5)}  &  $3.25 \pm 0.18$                   &     & $18.35 \pm 0.38$            & $2.43 \pm 0.10$    \\
        \shortstack[l]{Cutout}                          &  $3.13 \pm 0.12$                     &     & $20.16 \pm 0.35$            & $2.48 \pm 0.15$     \\
        \shortstack[l]{DropBlock}                       &  $4.18 \pm 0.07$                     &     & $22.36 \pm 0.15$            & $2.73 \pm 0.06$     \\
        \shortstack[l]{CutMix}                          &  $3.15 \pm 0.12$                     &     & $18.32 \pm 0.19$            & $2.43 \pm 0.05$     \\
        \shortstack[l]{Puzzle Mix}                       &   {\ul{$2.56 \pm 0.07$}}         &     &       $17.53 \pm 0.22$     & $2.43 \pm 0.07$   \\
        \shortstack[l]{\softpatchup{}}                  &  $2.61 \pm 0.05$   &     & {\ul{$16.73 \pm 0.11$}}  & {$\mathbf{2.08 \pm 0.07}$}     \\
        \shortstack[l]{\hardpatchup{}}                  &  {$\mathbf{2.53 \pm 0.07}$}    &     & {$\mathbf{16.13 \pm 0.19}$} & {\ul{$2.09 \pm 0.06$}}     \\ \bottomrule
        \end{tabular}}
\end{subtable}
 \captionsetup[subtable]{position = top}
    \captionsetup[table]{position=top}
    \caption{Image classification error rates on CIFAR10/100 and SVHN (five runs). The best performance result is shown in bold, the second-best is underlined. The lower is better.}
    \label{tab:tbl-cifar-svhn}
\end{table}

Since ManifoldMixup and Puzzle Mix show that they perform better than No-Mixup and Input Mixup on affine transformation and against adversarial attacks~\cite{manifold}, we exclude No-Mixup and Input Mixup for the tasks in the following sections.
Table~\ref{tab:imagenet_result} shows that \patchup{} achieves a better error rate compared to other methods in the ImageNet2012 dataset~\cite{russakovsky2015imagenet}. To have a fair comparison, we used the same experiment setup proposed in the CutMix paper (300 epochs). For \softpatchup{} we set the gamma, patchup\_block, and patchup\_prob to 0.6, 7, and 1.0, respectively. For Hard \patchup{} we set them to 0.5, 7, and 0.6, respectively.
\begin{table}[htbp!]
\setlength{\tabcolsep}{1pt}
\centering
    \begin{subtable}{.8\linewidth}
        \resizebox{\linewidth}{!}{
    \begin{tabular}{lllllll}
    \toprule
    \shortstack[l]{Method}      & \shortstack[l]{Top-1 Error (\%)}           & \shortstack[l]{Top-5 Error (\%)}         \\ \midrule
    \shortstack[l]{Vanilla$^{*}$}    & $23.68$ & $7.05$ \\
    \shortstack[l]{Input Mixup$^{*}$}    & $22.58$ & $6.40$ \\
    \shortstack[l]{Cutout$^{*}$} & $22.93$ & $6.66$ \\
    \shortstack[l]{ManifoldMixup$^{*}$} & $22.50$ & $6.21$ \\
    \shortstack[l]{FeatureCutMix$^{*}$}         & $21.80$ & $6.06$ \\
    \shortstack[l]{CutMix$^{*}$}         & $21.40$ & $5.92$ \\
    \shortstack[l]{PuzzleMix$^{**}$}         & $21.24$ & $5.71$ \\
    \shortstack[l]{\softpatchup{}}         & \ul{$21.06$} & \ul{$5.57$} \\
    \shortstack[l]{\hardpatchup{}}         & $\mathbf{20.75}$ & $\mathbf{5.29}$ \\\bottomrule
    \end{tabular}}
\end{subtable}
\caption{Classification error rates on on the ILSVRC2012 (ImageNet2012) dataset. We include results from~\cite{cutmix}$^*$ and~\cite{kim2020puzzle}$^{**}$. 
}
\label{tab:imagenet_result}
\end{table}
\begin{table*}[htbp!]
\setlength{\tabcolsep}{7pt}
\centering
    \begin{subtable}{.9\linewidth}
        \resizebox{\linewidth}{!}{
    \begin{tabular}{lllllll}
    \toprule
    \shortstack[l]{Transformation}      & \shortstack[l]{cutout}           & \shortstack[l]{\cutmix{}}          & \shortstack[l]{ManifoldMixup} & \shortstack[l]{Puzzle Mix}    & \shortstack[l]{\softpatchup{}}              & \shortstack[l]{\hardpatchup{}}             \\ \midrule
    \shortstack[l]{Rotate (-20, 20)}    & $37.45 \pm 0.53 $ & $35.42 \pm 0.33 $ & $35.44 \pm 0.57 $ & $31.70 \pm 0.66$ & \ul{$31.14 \pm 0.52$} & $\mathbf{30.41 \pm 0.52} $ \\
    \shortstack[l]{Rotate (-40, 40)}    & $58.75 \pm 0.10 $ & $57.83 \pm 0.59 $ & $54.42 \pm 0.95 $ & $54.04 \pm 0.80$ & \ul{$53.42 \pm 0.42$} & $\mathbf{49.96 \pm 0.80} $ \\
    \shortstack[l]{Shear (-28.6, 28.6)} & $36.55 \pm 0.49 $ & $34.15 \pm 0.47 $ & $34.15 \pm 0.42 $ & $31.63 \pm 0.69$ & $\mathbf{28.98 \pm 0.50} $ & \ul{$29.57 \pm 0.41$} \\
    \shortstack[l]{Shear (-57.3, 57.3)} & $57.74 \pm 0.57 $ & $53.64 \pm 0.59 $ & $55.44 \pm 0.68 $ & $52.49 \pm 0.39$ & $\mathbf{49.10 \pm 0.53} $ & \ul{$50.32 \pm 0.62$} \\
    \shortstack[l]{Scale (0.6)}         & $72.99 \pm 1.23 $ & $54.30 \pm 1.27 $ & $78.99 \pm 1.13 $ & $59.70 \pm 1.24$ & $\mathbf{46.25 \pm 1.20} $ & \ul{$50.06 \pm 2.69$} \\
    \shortstack[l]{Scale (0.8)}         & $35.09 \pm 0.86 $ & $29.38 \pm 0.58 $ & $34.62 \pm 0.37 $ & $30.37 \pm 0.36$ & $\mathbf{23.94 \pm 0.21} $ & \ul{$25.34 \pm 0.33$} \\
    \shortstack[l]{Scale (1.2)}         & $42.31 \pm 0.71 $ & $49.52 \pm 2.04 $ & \ul{$41.32 \pm 0.64$}  & $47.40 \pm 1.44$ & $43.41 \pm 0.65 $ & $\mathbf{38.01 \pm 0.70} $ \\
    \shortstack[l]{Scale (1.4)}         & $69.40 \pm 0.90 $ & $78.66 \pm 1.85 $ & $\mathbf{65.94 \pm 0.75} $ & $77.59 \pm 0.97$ & $77.07 \pm 1.19 $ & \ul{$66.34 \pm 1.22$} \\ \bottomrule
    \end{tabular}}
\end{subtable}
\caption{Error rates in the test set on samples subject to affine transformations for PreActResNet34 trained on CIFAR100 (repeated each test for five trained models). Best in bold, second best is underlined. The lower number is better.}
    \label{tab:affine_p34_result}
\end{table*}
\subsection{Robustness to Common Corruptions}
\begin{figure}[htbp!]
\centering
  \includegraphics[width=.9\linewidth]{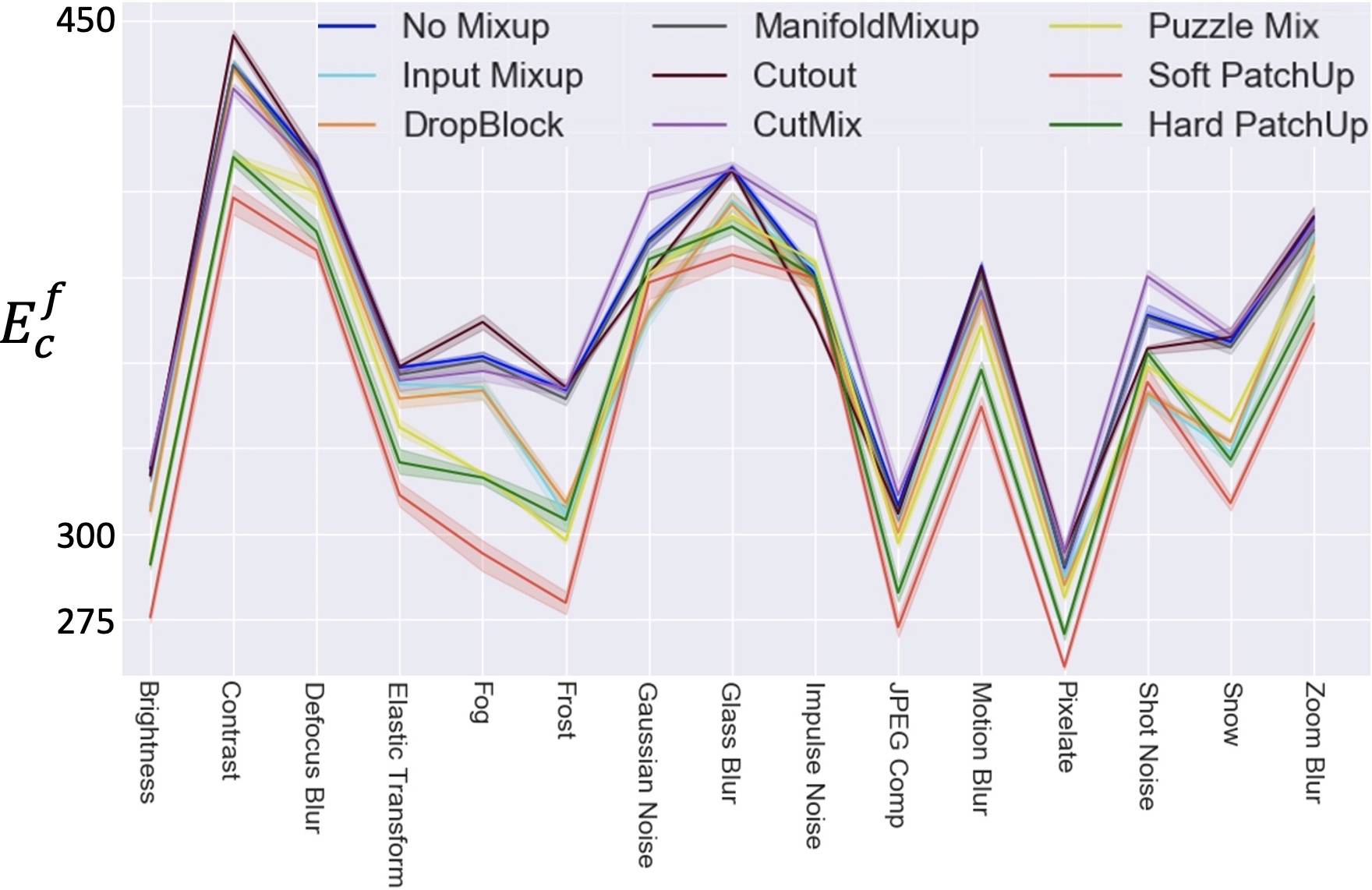}
	\caption{Errors in Tiny-ImageNet-C for WRN-28-10. The y-axis is the sum of error rates for each category. The x-axis represents the corruptions. The lower is better (five runs).}
\label{fig:tiny_c_cmp}
\end{figure}

The common corruption benchmark helps to evaluate the robustness of models against the input corruptions~\cite{hendrycks2019benchmarking}. It uses the 75 corruptions in 15 categories such that each has five levels of severity. We compare the methods robustness in Tiny-ImageNet-C for ResNet101/152, and WRN-28-10. So, we compute the sum of error denoted as $\mathrm{E}_{c}^{f}$ where $s$ is the level of severity and $c$ is corruption type such that $\mathrm{E}_{c}^{f}=\sum_{s=1}^{5} E_{s, c}^{f}$~\cite{hendrycks2019benchmarking}.
Fig.\ref{fig:tiny_c_cmp} shows \softpatchup{} leads the best performance in Tiny-ImageNet-C and \hardpatchup{} achieves the second-best. Figures 15a and b in Appendix show the comparison results in ResNet101 and 152.
\subsection{Generalization on Deformed Images}
\label{exp:deformed_images_test}
Affine transformations on the test set provide novel deformed data that can be used to evaluate the robustness of a method on out-of-distribution samples~\cite{manifold}. We trained PreActResNet34 and WRN-28-10 on the CIFAR100. Then, we created a deformed test from CIFAR100 by applying some affine transformations.
Table~\ref{tab:affine_p34_result} shows that \patchup{} provides the best performance on the affine transformed test and better generalization in PreActResNet34. Table 10 (F-appx.) shows that the quality of representations is improved by \patchup{} and it shows better generalization in the deformed test on WRN-28-10. Generalization is significantly improved by \patchup{} over existing methods by a large margin, as is the quality of representations.
\begin{figure}[htbp!]
\centering   
\begin{subfigure}{0.49\linewidth}
	\includegraphics[width=.95\linewidth]{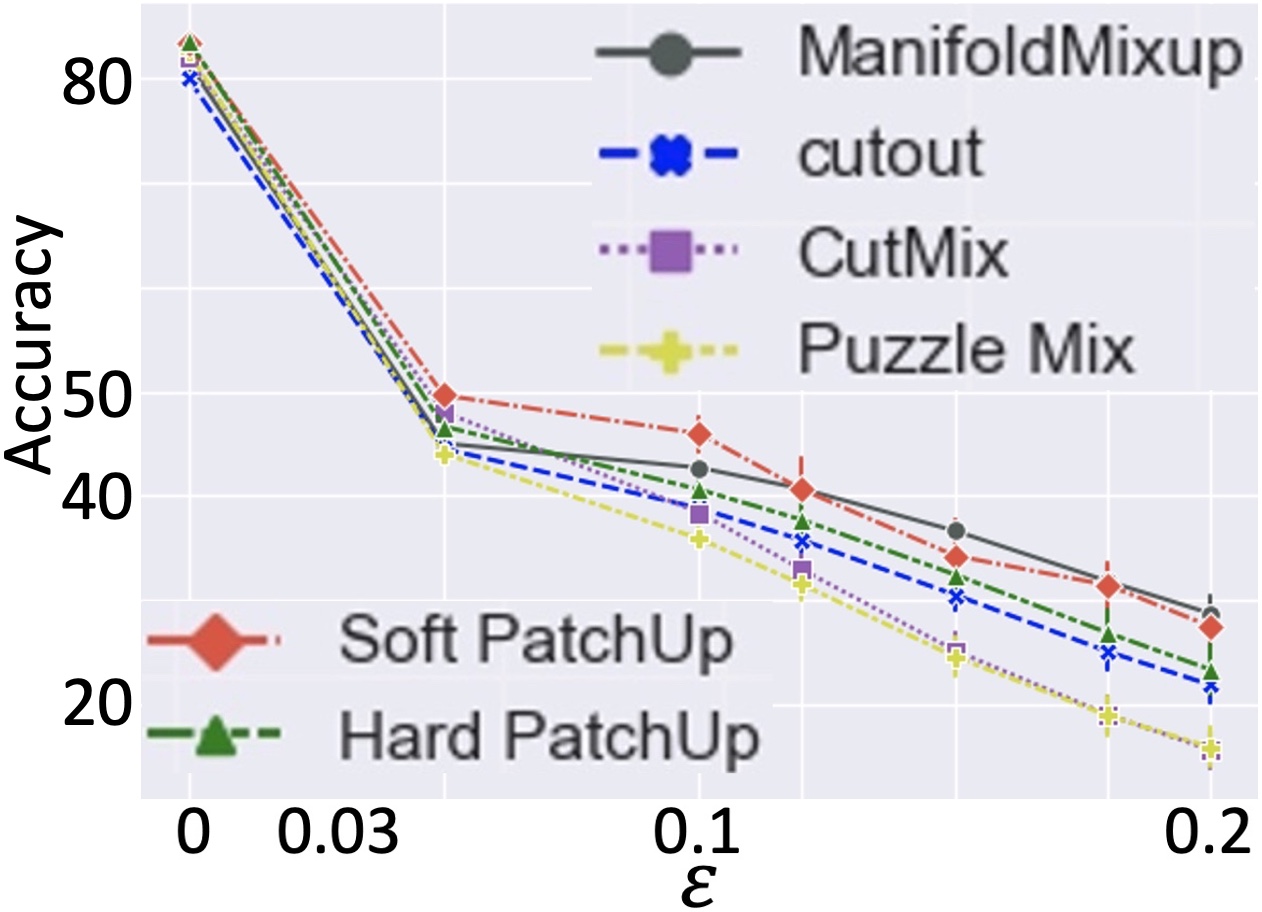} 
	\caption{CIFAR100.}
	\label{fig:attack_c100_p18}
	\end{subfigure}
\hfill
\begin{subfigure}{0.49\linewidth}
	\includegraphics[width=.95\linewidth]{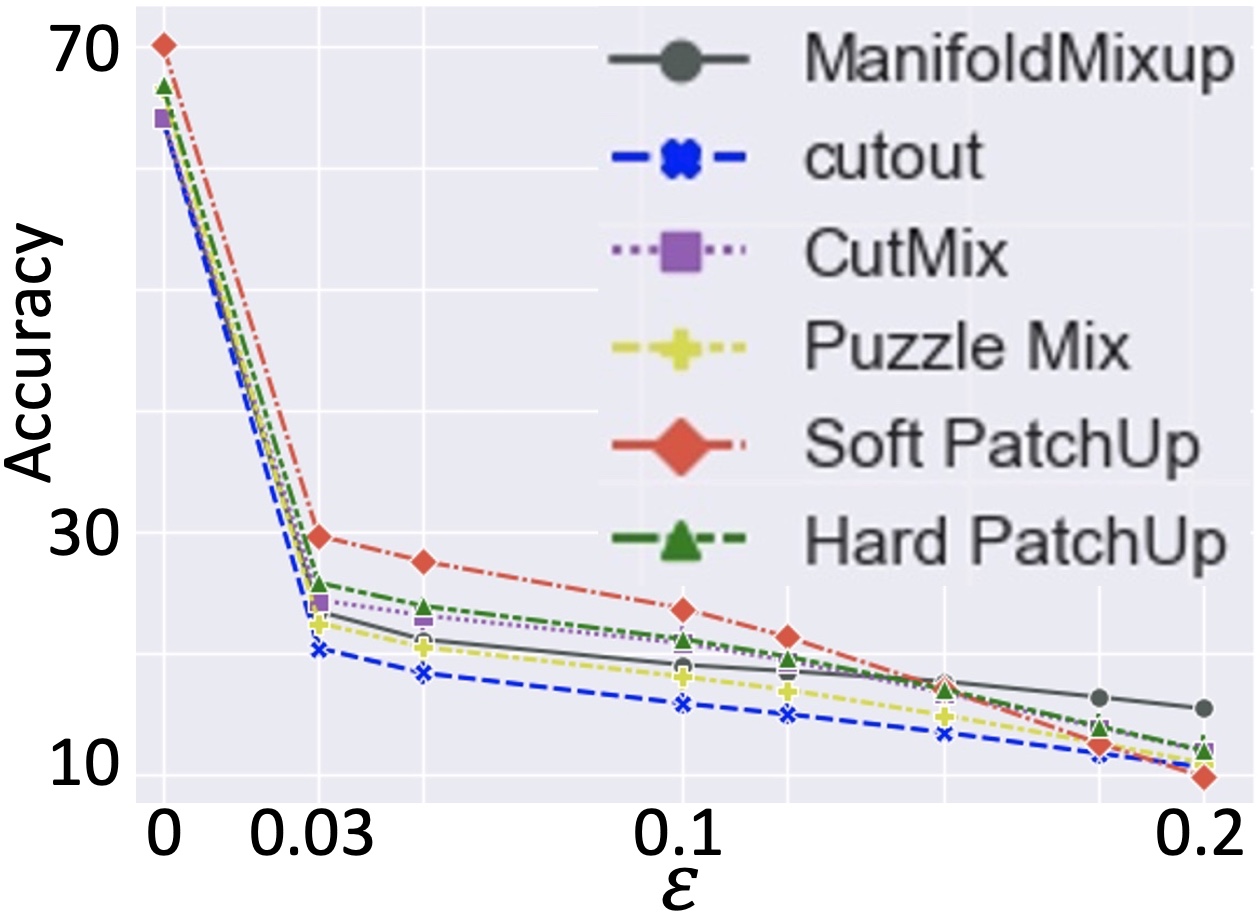} 
	\caption{Tiny-ImageNet.}
	\label{fig:attack_svhn_wrn}
	\end{subfigure}
\caption{WRN-28-10 robustness to FGSM attack (five runs). The y-axis is the accuracy against FGSM attack. The x-axis is the magnitude that controls the perturbation ($\epsilon$).}
\label{fig:adv_attack}
\end{figure}
\subsection{Robustness to Adversarial Examples}
Neural networks, trained with ERM, are often vulnerable to adversarial examples~\cite{szegedy_Zaremba}. Certain data-dependent methods can alleviate such fragility to adversarial examples by training the models with interpolated data. So, the robustness of a regularized model to adversarial examples can be considered as a criterion for comparison~\cite{mixup, manifold}. Fig.\ref{fig:adv_attack} compares the performance of the methods on CIFAR100 and Tiny-ImageNet with adversarial examples created by the FGSM attack described in~\cite{goodfellow2014FGSM}. Fig 13-appx. contains further comparison on  PreActResNet18/34 and WRN-28-10 for CIFAR10 and SVHN with FGSM attacks. 
Table~\ref{tab:tbl-tiny_imagenet_attacks} shows the robust accuracy (in the range of $[0,\,1]$) for the Foolbox benchmark~\cite{rauber2018foolbox} against the 7-steps DeepFool~\cite{moosavi2016deepfool}, Decoupled Direction and Norm (DDN)~\cite{Rony_2019_CVPR}, Carlini-Wagner (CW)~\cite{carlini2017towards}, and $\mathrm{PGD}_{L_{\infty}}$~\cite{madry2019deep} attacks with $\epsilon=\frac{8}{255}$.
We observe that \patchup{} is more robust to adversarial attacks compared to other methods. While \hardpatchup{} achieves better performance in terms of classification accuracy, \softpatchup{} seems to trade off a slight loss of accuracy in order to achieve more robustness.
\begin{table}[htbp!]
\setlength{\tabcolsep}{3pt} 
\centering
    \begin{subtable}{1.\linewidth}
        \resizebox{\linewidth}{!}{
        \begin{tabular}{lcccc}
        \toprule
        \shortstack[l]{Methods}                                   & \shortstack[l]{DeepFool}   & \shortstack[l]{DDN} & \shortstack[l]{CW} & \shortstack[l]{$\mathrm{PGD}_{L_{\infty}}$} \\ \midrule
        \shortstack[l]{No Mixup}                         & $0.17\pm 0.01$  & $0.18 \pm 0.01$   & $0.18\pm 0.01$   & $0.17\pm 0.01$  \\
        \shortstack[l]{Input Mixup}     & {\ul{$0.19\pm 0.01$}}  & $0.20 \pm 0.01$   & $0.20\pm 0.01$   & $0.19\pm 0.01$ \\
        \shortstack[l]{ManifoldMixup} & {$\mathbf{0.20\pm 0.01}$}  & {$\mathbf{0.20 \pm 0.01}$}   & {\ul{$0.20\pm 0.01$}}   & {\ul{$0.19\pm 0.01$}}  \\
        \shortstack[l]{Cutout}                           & $0.18\pm 0.01$  & $0.185 \pm 0.01$   & $0.19\pm 0.01$   & $0.18\pm 0.01$   \\
        \shortstack[l]{DropBlock}                        & $0.18\pm 0.01$  & $0.186 \pm 0.01$   & $0.19\pm 0.01$   & $0.18\pm 0.01$     \\
        \shortstack[l]{CutMix}                           & $0.17\pm 0.01$  & $0.171 \pm 0.01$   & $0.17\pm 0.01$   & $0.17\pm 0.01 $ \\ 
        \shortstack[l]{Puzzle Mix}                       & $0.19\pm 0.01$  & $0.191 \pm 0.01$   & $0.19\pm 0.01$   & $0.19\pm 0.01$   \\
        \shortstack[l]{\softpatchup{}}                   & $0.19\pm 0.01$  & {\ul{$0.20 \pm 0.01$}}   & {$\mathbf{0.20\pm 0.01}$}   & {$\mathbf{0.19\pm 0.01}$}  \\
        \shortstack[l]{\hardpatchup{}}                   & $0.18\pm 0.01$  & $0.19 \pm 0.01$   & $0.19\pm 0.01$   & $0.18\pm 0.01$   \\ \bottomrule
        \end{tabular}}
\end{subtable}
\captionsetup[subtable]{position = top}
    \captionsetup[table]{position=top}
    \caption{Robust Accuracy of WRN-28-10 in the Tiny-ImageNet dataset against adversarial 7-steps attacks with $\epsilon=\frac{8}{255}$. The $\alpha$ is $0.2$ for Mixup, ManifoldMixup, and Puzzle Mix. The best performance result is shown in bold, second best is underlined. The higher number is better (five runs).}
    \label{tab:tbl-tiny_imagenet_attacks}
\end{table}

\subsection{Effect on Activations}
\begin{figure}[hbp!]
\centering
	\includegraphics[width=1.\linewidth]{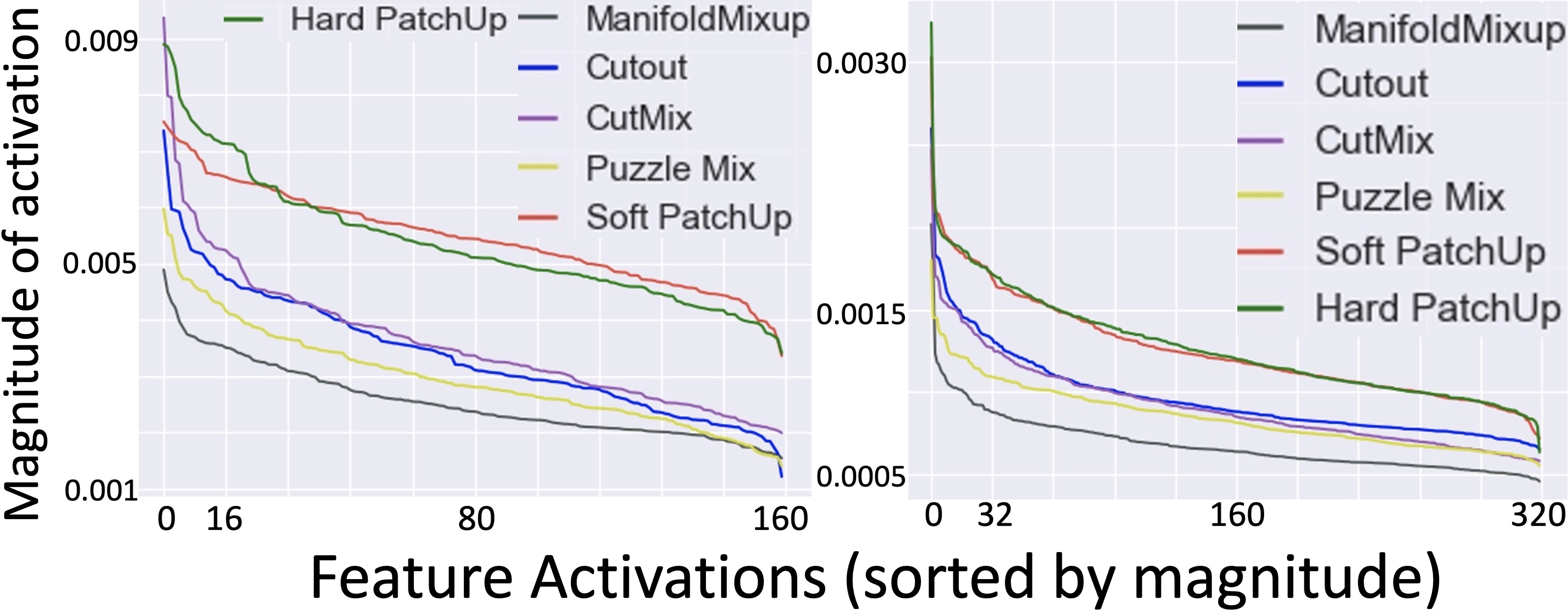}
	\caption{1st (left) and 2st (right) Residual Block.}
	\label{fig:activation_layer12}
\caption{The effect of the methods on activations in WRN-28-10 for CIFAR100 test set on the first and second Residual Blocks. Each curve is the magnitude of the feature activations, sorted in descending value, and averaged over all test samples. The higher magnitude indicates a wider variety of the produced features by the model at each block.}
\label{fig:activation_cmp}
\end{figure}

To study the effect of the methods on the activations in the residual blocks, we compared the mean magnitude of feature activations in the residual blocks following~\cite{cutout} in WRN-28-10 for CIFAR100 test set. We train the models with each method and then calculate the magnitudes of activations in the test set. The higher mean magnitude of features shows that the models tried to produce a wider variety of features in the residual blocks~\cite{cutout}. Our WRN-28-10 has a conv2d module followed by three residual blocks. We selected $k$ randomly such that $k\in \{0, 1, 2, 3\}$. And, we apply the ManifoldMixup and \patchup{} in either input space, the first conv2d, and the first or second residual blocks (results in Fig.\ref{fig:activation_cmp}).
Figure~\ref{fig:activation_layer12} shows that \patchup{} produces more diverse features in the layers where we apply \patchup{}. Fig. 14-appx. shows the results in the first conv2d, first residual, second residual, and third residual blocks. Since we are not applying the \patchup{} in the third residual block, the mean magnitude of the feature activations are below, but very close to, Cutout and \cutmix{}. This also shows that producing a wide variety of features can be an advantage for a model. But, according to our experiments, a larger magnitude of activations does not always lead to better performance. Fig.\ref{fig:activation_cmp} shows that for ManifoldMixup, the mean magnitude of the feature activations is less than others. But, it performs better than Cutout and \cutmix{} in most tasks. 
\subsection{Significance of loss terms and analysis of \textbf{\textit{k}}}
\label{sec:patchup_loss_k}
\patchup{} uses the loss that is introduced in Equation~\ref{eq:swap_loss}. We can paraphrase the \patchup{} learning objective for this ablation study as follows:

\begin{align}
\begin{array}{l}
L(f)=\mathbb{E}_{\left(x_{i}, y_{i}\right)_{\sim P}} \mathbb{E}_{\left(x_{j}, y_{j}\right)_{\sim P}} \mathbb{E}_{\lambda \sim \operatorname{Beta}(\alpha, \alpha)} \mathbb{E}_{k \sim \mathcal{S}}\left(L_{1}+L_{2}\right)
\end{array}
\end{align}
where $L_{1} = \text{Mix}_{p_{u}} [ \ell(f_{k}(\phi_k), y_i), \ell(f_{k}(\phi_k), Y)]$ and $L_{2} = \ell(f_{k}(\phi_k), W(y_i, y_j))$. We also show the effect of $L_{1}$ and $L_{2}$ in \patchup{} loss. Table~\ref{tab:loss_study} shows the error rate on the validation set for WRN-28-10 on CIFAR100. This shows the summation of the $L_{1}$ and $L_{2}$ reduces error rate by $.1\%$ in \patchup{}. We conducted an experiment to show the importance of random layer selection in PatchUp. Table~\ref{tab:layer_study_k} shows the contribution of the random selection of the layer in the overall performance of the method. In the left-most column 1/2/3 refers to PatchUp being applied to only one layer (more details in section ``B'' in appx.). 

\begin{figure}[htbp!]
\centering
\includegraphics[width=.85\linewidth]{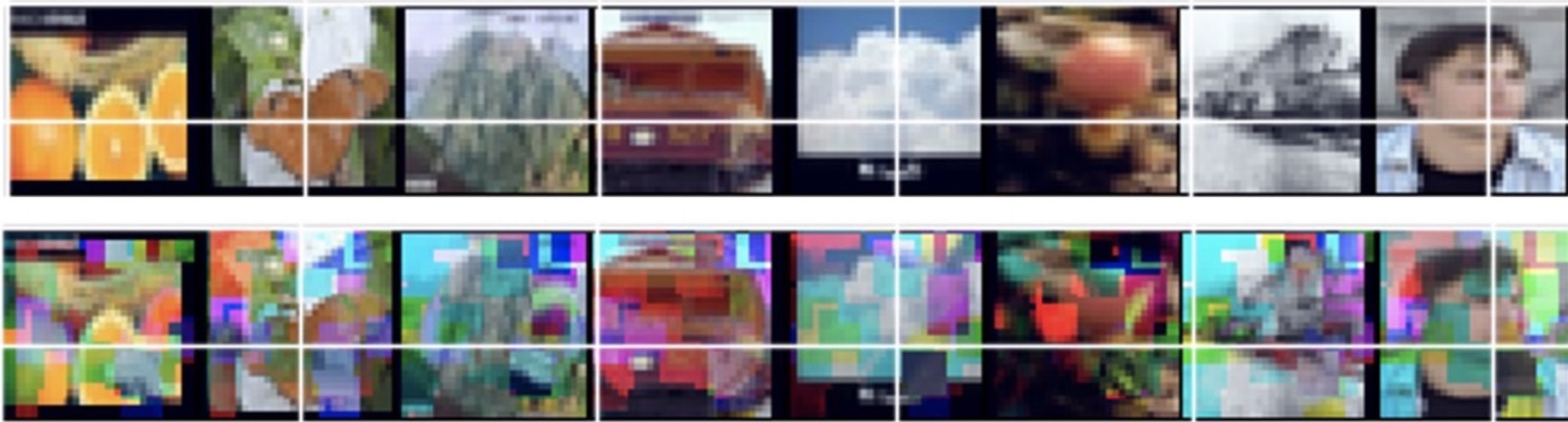} 
	\caption{(top) Original samples. (bottom) Hard PatchUp output using PatchUp Binary Mask on input images.}
	\label{fig:layer_zoro}
\end{figure}

As noted in section~``\ref{sec:patchup_in_input}'', the PatchUp mask is ``too strong'' for the input space. Fig.\ref{fig:layer_zoro} shows that the PatchUp mask often drastically destroys the semantic concepts in the input images. Thus, we select one random rectangular patch in the input space (similar to CutMix). However, the learning objective in (k = 0) is still the PatchUp objective that is different from CutMix. The last row in table~\ref{tab:layer_study_k} shows the negative effect of applying the PatchUp mask in the input space.
\begin{table}[htbp!]
{\renewcommand{\arraystretch}{1.}
\setlength{\tabcolsep}{3pt}
\centering
    \begin{subtable}{.9\linewidth}
        \resizebox{\linewidth}{!}{
        \begin{tabular}{llll}
        \toprule
        Simple & \multicolumn{3}{c}{Error Rate:} \\
        WRN-28-10 & \multicolumn{3}{c}{$23.26 \pm 0.59$} \\
        \midrule
        {}     & Error with $L1$ & Error with $L2$ & Error with $L (f)$ \\ \cmidrule(lr){2-4}
        \softpatchup & $16.86 \pm 0.67$   & $16.87 \pm 0.34$   &  $\mathbf{16.75 \pm 0.29}$\\
        \cmidrule(lr){2-4}
        \hardpatchup & $16.14 \pm 0.23$   & $16.79 \pm 0.46$   &  $\mathbf{16.02 \pm 0.36}$ \\
        \bottomrule
        \end{tabular}}
    \end{subtable}
    \captionsetup[subtable]{position = top}
    \captionsetup[table]{position=top}
    \caption{The validation error on CIFAR100 for WRN-28-10 with \textit{Hard and Soft PatchUp}. The lower is better (five runs).}
    \label{tab:loss_study}
    }
\end{table}
\begin{table}[htbp!]
{\renewcommand{\arraystretch}{1.}
\setlength{\tabcolsep}{5pt}
\centering
    \begin{subtable}{.85\linewidth}
        \resizebox{\linewidth}{!}{
        \begin{tabular}{l llll }
        \toprule
        {\textbf{layer}} & {Val Error}  & {Test Error} & {Test NLL} \\
        \midrule
		$1$ & $18.43 \pm 0.44$  & $17.86 \pm 0.16$ & $0.73 \pm 0.01$ \\
		\cmidrule(lr){2-5}
		$2$ & $22.54 \pm 0.80$  & $21.42 \pm 0.28$ & $0.85 \pm 0.01$ \\
		\cmidrule(lr){2-5}
		$3$ & $26.17 \pm 0.50$ & $25.25 \pm 0.14$ & $1.14 \pm 0.03$ \\
        \midrule
        \multicolumn{2}{l}{\shortstack[l]{Random selection}} & \\
         & $\mathbf{16.38 \pm 0.47}$  & {$\mathbf{16.13 \pm 0.20}$}              &  $\mathbf{0.66 \pm 0.02}$ \\
        \midrule
        \multicolumn{2}{l}{\shortstack[l]{PatchUp Masks in }} & \\
         $\mathbf{k=0}$ &$16.98 \pm 0.34$ & $16.90 \pm 0.0.21$ & $0.67 \pm 0.01$\\
        \bottomrule
        \end{tabular}}
    \end{subtable}
  \captionsetup[subtable]{position = top}
    \captionsetup[table]{position=top}
    \caption{WRN-28-10 using \hardpatchup{} on CIFAR100 (five runs).}
    \label{tab:layer_study_k}}
\end{table}
\section{Conclusion}
\label{sec:conclusion}
We presented \patchup{}, a simple and efficient regularizer scheme for CNNs that alleviates some of the drawbacks of the previous mixing-based regularizers. Our experimental results show that with the proposed approach, \patchup{}, we can achieve state-of-the-art results on image classification tasks across different architectures and datasets. Similar to previous mixing-based approaches, our approach also has the advantage of avoiding any added computational overhead. The strong test accuracy achieved by \patchup{}, with no additional computational overhead, makes it particularly appealing for practical applications. 
\section{Acknowledgments}
\label{sec:acknowledgments}
We would like to acknowledge Compute Canada and Calcul Quebec for providing the computing resources used in this work. The authors would also like to thank Damien Scieur, Hannah Alsdurf, Alexia Jolicoeur-Martineau, and Yassine Yaakoubi for reviewing the manuscript. SC is supported by a Canada CIFAR AI Chair and an NSERC Discovery Grant.


\medskip
\bibliography{aaai22}

\begin{thebibliography}{41}
\providecommand{\natexlab}[1]{#1}

\bibitem[{{Achille} and {Soatto}(2018)}]{information_dropout}
{Achille}, A.; and {Soatto}, S. 2018.
\newblock Information Dropout: Learning Optimal Representations Through Noisy
  Computation.
\newblock \emph{IEEE Transactions on Pattern Analysis and Machine
  Intelligence}, 40(12): 2897--2905.

\bibitem[{Arpit et~al.(2017)Arpit, Jastrz{e}bski, Ballas, Krueger, Bengio,
  Kanwal, Maharaj, Fischer, Courville, Bengio et~al.}]{arpit2017closer}
Arpit, D.; Jastrz{e}bski, S.; Ballas, N.; Krueger, D.; Bengio, E.; Kanwal,
  M.~S.; Maharaj, T.; Fischer, A.; Courville, A.; Bengio, Y.; et~al. 2017.
\newblock A closer look at memorization in deep networks.
\newblock In \emph{Proceedings of the 34th International Conference on Machine
  Learning-Volume 70}, 233--242. JMLR. org.

\bibitem[{Carlini and Wagner(2017)}]{carlini2017towards}
Carlini, N.; and Wagner, D. 2017.
\newblock Towards evaluating the robustness of neural networks.
\newblock In \emph{2017 ieee symposium on security and privacy (sp)}, 39--57.
  IEEE.

\bibitem[{Chrabaszcz, Loshchilov, and Hutter(2017)}]{chrabaszcz2017downsampled}
Chrabaszcz, P.; Loshchilov, I.; and Hutter, F. 2017.
\newblock A Downsampled Variant of ImageNet as an Alternative to the CIFAR
  datasets.
\newblock arXiv:1707.08819.

\bibitem[{Cubuk et~al.(2019)Cubuk, Zoph, Mane, Vasudevan, and
  Le}]{cubuk2019autoaugment}
Cubuk, E.~D.; Zoph, B.; Mane, D.; Vasudevan, V.; and Le, Q.~V. 2019.
\newblock AutoAugment: Learning Augmentation Policies from Data.
\newblock arXiv:1805.09501.

\bibitem[{DeVries and Taylor(2017)}]{cutout}
DeVries, T.; and Taylor, G.~W. 2017.
\newblock Improved Regularization of Convolutional Neural Networks with Cutout.
\newblock arXiv:1708.04552.

\bibitem[{Gal and Ghahramani(2016)}]{gal2016theoretically}
Gal, Y.; and Ghahramani, Z. 2016.
\newblock A theoretically grounded application of dropout in recurrent neural
  networks.
\newblock In \emph{Advances in neural information processing systems},
  1019--1027.

\bibitem[{Ghiasi, Lin, and Le(2018)}]{dropblock}
Ghiasi, G.; Lin, T.; and Le, Q.~V. 2018.
\newblock DropBlock: {A} regularization method for convolutional networks.
\newblock \emph{CoRR}, abs/1810.12890.

\bibitem[{Goodfellow, Bengio, and Courville(2016)}]{goodfellow2016deep}
Goodfellow, I.; Bengio, Y.; and Courville, A. 2016.
\newblock \emph{Deep learning}.
\newblock MIT press.

\bibitem[{Goodfellow, Shlens, and Szegedy(2014)}]{goodfellow2014FGSM}
Goodfellow, I.~J.; Shlens, J.; and Szegedy, C. 2014.
\newblock Explaining and Harnessing Adversarial Examples.
\newblock arXiv:1412.6572.

\bibitem[{Guo, Mao, and Zhang(2018)}]{DBLP:Hongyu}
Guo, H.; Mao, Y.; and Zhang, R. 2018.
\newblock MixUp as Locally Linear Out-Of-Manifold Regularization.
\newblock \emph{CoRR}, abs/1809.02499.

\bibitem[{He et~al.(2015)He, Zhang, Ren, and Sun}]{Kaiming}
He, K.; Zhang, X.; Ren, S.; and Sun, J. 2015.
\newblock Deep Residual Learning for Image Recognition.
\newblock \emph{CoRR}, abs/1512.03385.

\bibitem[{He et~al.(2016)He, Zhang, Ren, and Sun}]{he2016identity}
He, K.; Zhang, X.; Ren, S.; and Sun, J. 2016.
\newblock Identity Mappings in Deep Residual Networks.
\newblock arXiv:1603.05027.

\bibitem[{Hendrycks and Dietterich(2019)}]{hendrycks2019benchmarking}
Hendrycks, D.; and Dietterich, T. 2019.
\newblock Benchmarking Neural Network Robustness to Common Corruptions and
  Perturbations.
\newblock arXiv:1903.12261.

\bibitem[{Hendrycks et~al.(2020)Hendrycks, Mu, Cubuk, Zoph, Gilmer, and
  Lakshminarayanan}]{hendrycks2020augmix}
Hendrycks, D.; Mu, N.; Cubuk, E.~D.; Zoph, B.; Gilmer, J.; and
  Lakshminarayanan, B. 2020.
\newblock AugMix: A Simple Data Processing Method to Improve Robustness and
  Uncertainty.
\newblock arXiv:1912.02781.

\bibitem[{{Hinton} et~al.(2012){Hinton}, {Deng}, {Yu}, {Dahl}, {Mohamed},
  {Jaitly}, {Senior}, {Vanhoucke}, {Nguyen}, {Sainath}, and
  {Kingsbury}}]{hinton}
{Hinton}, G.; {Deng}, L.; {Yu}, D.; {Dahl}, G.~E.; {Mohamed}, A.; {Jaitly}, N.;
  {Senior}, A.; {Vanhoucke}, V.; {Nguyen}, P.; {Sainath}, T.~N.; and
  {Kingsbury}, B. 2012.
\newblock Deep Neural Networks for Acoustic Modeling in Speech Recognition: The
  Shared Views of Four Research Groups.
\newblock \emph{IEEE Signal Processing Magazine}, 29(6): 82--97.

\bibitem[{Kim, Choo, and Song(2020)}]{kim2020puzzle}
Kim, J.-H.; Choo, W.; and Song, H.~O. 2020.
\newblock Puzzle Mix: Exploiting Saliency and Local Statistics for Optimal
  Mixup.
\newblock arXiv:2009.06962.

\bibitem[{Kingma, Salimans, and Welling(2015)}]{kingma2015variational}
Kingma, D.~P.; Salimans, T.; and Welling, M. 2015.
\newblock Variational dropout and the local reparameterization trick.
\newblock \emph{Advances in neural information processing systems}, 28:
  2575--2583.

\bibitem[{Kingma and Welling(2013)}]{kingma2013auto}
Kingma, D.~P.; and Welling, M. 2013.
\newblock Auto-encoding variational bayes.
\newblock \emph{arXiv preprint arXiv:1312.6114}.

\bibitem[{Krizhevsky(2009)}]{dataset:cifar}
Krizhevsky, A. 2009.
\newblock Learning Multiple Layers of Features from Tiny Images.

\bibitem[{Krizhevsky, Sutskever, and Hinton(2012)}]{krizhevsky}
Krizhevsky, A.; Sutskever, I.; and Hinton, G.~E. 2012.
\newblock ImageNet Classification with Deep Convolutional Neural Networks.
\newblock In Pereira, F.; Burges, C. J.~C.; Bottou, L.; and Weinberger, K.~Q.,
  eds., \emph{Advances in Neural Information Processing Systems 25},
  1097--1105. Curran Associates, Inc.

\bibitem[{Krueger et~al.(2016)Krueger, Maharaj, Kram{\'a}r, Pezeshki, Ballas,
  Ke, Goyal, Bengio, Courville, and Pal}]{krueger2016zoneout}
Krueger, D.; Maharaj, T.; Kram{\'a}r, J.; Pezeshki, M.; Ballas, N.; Ke, N.~R.;
  Goyal, A.; Bengio, Y.; Courville, A.; and Pal, C. 2016.
\newblock Zoneout: Regularizing rnns by randomly preserving hidden activations.
\newblock \emph{arXiv preprint arXiv:1606.01305}.

\bibitem[{Madry et~al.(2019)Madry, Makelov, Schmidt, Tsipras, and
  Vladu}]{madry2019deep}
Madry, A.; Makelov, A.; Schmidt, L.; Tsipras, D.; and Vladu, A. 2019.
\newblock Towards Deep Learning Models Resistant to Adversarial Attacks.
\newblock arXiv:1706.06083.

\bibitem[{Mai et~al.(2019)Mai, Hu, Chen, Shen, and Shen}]{mai2019metamixup}
Mai, Z.; Hu, G.; Chen, D.; Shen, F.; and Shen, H.~T. 2019.
\newblock MetaMixUp: Learning Adaptive Interpolation Policy of MixUp with
  Meta-Learning.
\newblock arXiv:1908.10059.

\bibitem[{Moosavi-Dezfooli, Fawzi, and Frossard(2016)}]{moosavi2016deepfool}
Moosavi-Dezfooli, S.-M.; Fawzi, A.; and Frossard, P. 2016.
\newblock Deepfool: a simple and accurate method to fool deep neural networks.
\newblock In \emph{Proceedings of the IEEE conference on computer vision and
  pattern recognition}, 2574--2582.

\bibitem[{Netzer et~al.(2011)Netzer, Wang, Coates, Bissacco, Wu, and
  Ng}]{netzer2011reading}
Netzer, Y.; Wang, T.; Coates, A.; Bissacco, A.; Wu, B.; and Ng, A.~Y. 2011.
\newblock Reading Digits in Natural Images with Unsupervised Feature Learning.
\newblock In \emph{NIPS Workshop on Deep Learning and Unsupervised Feature
  Learning 2011}.

\bibitem[{Rauber, Brendel, and Bethge(2018)}]{rauber2018foolbox}
Rauber, J.; Brendel, W.; and Bethge, M. 2018.
\newblock Foolbox: A Python toolbox to benchmark the robustness of machine
  learning models.
\newblock arXiv:1707.04131.

\bibitem[{Ren et~al.(2015)Ren, He, Girshick, and Sun}]{NIPS2015_Ren}
Ren, S.; He, K.; Girshick, R.; and Sun, J. 2015.
\newblock Faster R-CNN: Towards Real-Time Object Detection with Region Proposal
  Networks.
\newblock In Cortes, C.; Lawrence, N.~D.; Lee, D.~D.; Sugiyama, M.; and
  Garnett, R., eds., \emph{Advances in Neural Information Processing Systems
  28}, 91--99. Curran Associates, Inc.

\bibitem[{Rony et~al.(2019)Rony, Hafemann, Oliveira, Ayed, Sabourin, and
  Granger}]{Rony_2019_CVPR}
Rony, J.; Hafemann, L.~G.; Oliveira, L.~S.; Ayed, I.~B.; Sabourin, R.; and
  Granger, E. 2019.
\newblock Decoupling Direction and Norm for Efficient Gradient-Based L2
  Adversarial Attacks and Defenses.
\newblock In \emph{Proceedings of the IEEE/CVF Conference on Computer Vision
  and Pattern Recognition (CVPR)}.

\bibitem[{Russakovsky et~al.(2015)Russakovsky, Deng, Su, Krause, Satheesh, Ma,
  Huang, Karpathy, Khosla, Bernstein, Berg, and
  Fei-Fei}]{russakovsky2015imagenet}
Russakovsky, O.; Deng, J.; Su, H.; Krause, J.; Satheesh, S.; Ma, S.; Huang, Z.;
  Karpathy, A.; Khosla, A.; Bernstein, M.; Berg, A.~C.; and Fei-Fei, L. 2015.
\newblock ImageNet Large Scale Visual Recognition Challenge.
\newblock arXiv:1409.0575.

\bibitem[{Srivastava et~al.(2014)Srivastava, Hinton, Krizhevsky, Sutskever, and
  Salakhutdinov}]{JMLR:v15:srivastava14a}
Srivastava, N.; Hinton, G.; Krizhevsky, A.; Sutskever, I.; and Salakhutdinov,
  R. 2014.
\newblock Dropout: A Simple Way to Prevent Neural Networks from Overfitting.
\newblock \emph{Journal of Machine Learning Research}, 15: 1929--1958.

\bibitem[{Sutskever, Vinyals, and Le(2014)}]{seq2seq}
Sutskever, I.; Vinyals, O.; and Le, Q.~V. 2014.
\newblock Sequence to Sequence Learning with Neural Networks.
\newblock In \emph{Proceedings of the 27th International Conference on Neural
  Information Processing Systems - Volume 2}, NIPS’14, 3104–3112.
  Cambridge, MA, USA: MIT Press.

\bibitem[{Szegedy et~al.(2013)Szegedy, Zaremba, Sutskever, Bruna, Erhan,
  Goodfellow, and Fergus}]{szegedy_Zaremba}
Szegedy, C.; Zaremba, W.; Sutskever, I.; Bruna, J.; Erhan, D.; Goodfellow, I.;
  and Fergus, R. 2013.
\newblock Intriguing properties of neural networks.
\newblock \emph{arXiv preprint arXiv:1312.6199}.

\bibitem[{Tishby and Zaslavsky(2015)}]{DBLP:Tishby}
Tishby, N.; and Zaslavsky, N. 2015.
\newblock Deep Learning and the Information Bottleneck Principle.
\newblock \emph{CoRR}, abs/1503.02406.

\bibitem[{Tompson et~al.(2014{\natexlab{a}})Tompson, Goroshin, Jain, LeCun, and
  Bregler}]{DBLP:spatial_dropout}
Tompson, J.; Goroshin, R.; Jain, A.; LeCun, Y.; and Bregler, C.
  2014{\natexlab{a}}.
\newblock Efficient Object Localization Using Convolutional Networks.
\newblock \emph{CoRR}, abs/1411.4280.

\bibitem[{Tompson et~al.(2014{\natexlab{b}})Tompson, Goroshin, Jain, LeCun, and
  Bregler}]{jonathan}
Tompson, J.; Goroshin, R.; Jain, A.; LeCun, Y.; and Bregler, C.
  2014{\natexlab{b}}.
\newblock Efficient Object Localization Using Convolutional Networks.
\newblock \emph{CoRR}, abs/1411.4280.

\bibitem[{Vaswani et~al.(2017)Vaswani, Shazeer, Parmar, Uszkoreit, Jones,
  Gomez, Kaiser, and Polosukhin}]{transformers}
Vaswani, A.; Shazeer, N.; Parmar, N.; Uszkoreit, J.; Jones, L.; Gomez, A.~N.;
  Kaiser, L.; and Polosukhin, I. 2017.
\newblock Attention Is All You Need.
\newblock \emph{CoRR}, abs/1706.03762.

\bibitem[{Verma et~al.(2019)Verma, Lamb, Beckham, Najafi, Mitliagkas,
  Lopez-Paz, and Bengio}]{manifold}
Verma, V.; Lamb, A.; Beckham, C.; Najafi, A.; Mitliagkas, I.; Lopez-Paz, D.;
  and Bengio, Y. 2019.
\newblock Manifold Mixup: Better Representations by Interpolating Hidden
  States.
\newblock In Chaudhuri, K.; and Salakhutdinov, R., eds., \emph{Proceedings of
  the 36th International Conference on Machine Learning}, volume~97 of
  \emph{Proceedings of Machine Learning Research}, 6438--6447. Long Beach,
  California, USA: PMLR.

\bibitem[{Yun et~al.(2019)Yun, Han, Oh, Chun, Choe, and Yoo}]{cutmix}
Yun, S.; Han, D.; Oh, S.~J.; Chun, S.; Choe, J.; and Yoo, Y. 2019.
\newblock CutMix: Regularization Strategy to Train Strong Classifiers with
  Localizable Features.
\newblock arXiv:1905.04899.

\bibitem[{Zagoruyko and Komodakis(2017)}]{zagoruyko2017wide}
Zagoruyko, S.; and Komodakis, N. 2017.
\newblock Wide Residual Networks.
\newblock arXiv:1605.07146.

\bibitem[{Zhang et~al.(2017)Zhang, Cisse, Dauphin, and Lopez-Paz}]{mixup}
Zhang, H.; Cisse, M.; Dauphin, Y.~N.; and Lopez-Paz, D. 2017.
\newblock mixup: Beyond Empirical Risk Minimization.
\newblock arXiv:1710.09412.

\end{thebibliography}

\appendix

\onecolumn
\addcontentsline{toc}{section}{Appendices}
\section*{Appendices}
\section{Algorithm}
\label{appendix:algorithm}
In this appendix, we provide a detailed algorithm for implementing \patchup{}. As with most regularization techniques, \patchup{} also has two modes (either inference or training). It also needs the combining type (either \softpatchup{} or \hardpatchup{}), $\gamma$, and $block\_size$. This algorithm shows how \patchup{} generates a new hidden representation from $(g_{k}(x_{i}), y_{i})$ and $(g_{k}(x_{j}), y_{j})$. Lines 4 to 9 in the algorithm\ref{alg:patchup} are the binary mask creation process used in both \softpatchup{} and \hardpatchup{}.
\begin{algorithm}[tbh]
\setstretch{1.}
   \caption{\patchup}
   \label{alg:patchup}
   \hspace*{.5cm} \textbf{Input:}\\
   \hspace*{1.5cm} $(g_{k}(x_{i}), y_{i})$: the hidden representation for the sample $(x_{i},\,y_{i})$ at layer $k$.\\
   \hspace*{1.5cm} $(g_{k}(x_{j}), y_{j})$: the hidden representation for the sample $(x_{j},\,y_{j})$ at layer $k$.\\
   \hspace*{1.5cm} $mode:$ \text{either \textit{inference} or \textit{training}.}\\
   \hspace*{1.5cm} ${mixing\_type}$: \textit{soft} or \textit{hard}.\\
   \hspace*{1.5cm} ${\gamma}$: the probability of altering a feature.\\
   \hspace*{1.5cm} ${block\_size}$: the size of each block in the binary mask.\\
    \hspace*{.5cm} \textbf{Output} \\
    \hspace*{1.5cm} $y_{i}$, $y_{j}$: original labels for samples $i$ and $j$.\\
    \hspace*{1.5cm} $H'$: the new hidden representation computed by \patchup{}.\\
    \hspace*{1.5cm} $p_{u}$: The portion of the feature maps that remained unchanged.\\
    \hspace*{1.5cm} $Y$: the target corresponding to the changed features.\\
    \hspace*{1.5cm} $W$: re-weighted target according to the interpolation policy.\\
\begin{algorithmic}[1]
    \IF {${mode}$ == \textit{Inference}}
        \STATE {return $(g_{k}(x_{i}), y_{i})$, $(g_{k}(x_{j}), y_{j})$}
    \ENDIF
    \STATE $kernel\_size \gets (block\_size,\,block\_size$)
    \STATE $stride \gets (1,\,1)$
    \STATE $padding \gets (\frac{block\_size}{2}, \frac{block\_size}{2})$
    \STATE $\gamma_{adj} \gets$ adjust $\gamma$ using (\ref{equ:gamma})
    \STATE $holes \gets max\_pool2d(Bernoulli(\gamma_{adj}),\;kernel\_size,\; stride,\;$ $padding)$
    \STATE $Mask \gets 1 - holes$
    \STATE $ unchanged \gets Mask \odot g_{k}(x_{i})$
    \STATE $p_{u} \gets \text{calculate the portion of changed features map.}$ 
    \STATE $ Patch_{i} \gets holes \odot g_{k}(x_{i})$
    \STATE $ Patch_{j} \gets holes \odot g_{k}(x_{j})$
    \IF {${mixing\_type}$ == \textit{hard}}
        \STATE $ Patch_{i} \gets Patch_{j} $
        \STATE $ Y \gets y_{j}$
        \STATE $W$ $\gets$ $W_{hard}(y_{i},\;y_{j})$  using (\ref{eq:target_reweighted})
    \ELSIF{${mixing\_type}$ == \textit{soft}}
        \STATE $\lambda \sim Beta(\alpha,\;\alpha)$ 
        \STATE $ Y \gets \textit{Mix}_{\lambda}(y_i,\;y_j)$
        \STATE $W$ $\gets$ $W_{soft}(y_{i},\;y_{j})$  using (\ref{eq:target_reweighted})
        \STATE $ Patch_{i} \gets \textit{Mix}_{\lambda}(Patch_{i},\;Patch_{j})$
    \ENDIF
    \STATE $ H' \gets unchanged + Patch_{i}$
    \STATE {{\bfseries return} {$y_{i},\,y_{j},\,H',\,p_{u},\,Y,\,W$}}

\end{algorithmic}
\end{algorithm}

\begin{figure}[htbp]
  \centering
\begin{subfigure}{0.47\linewidth}
	\label{fig:mixer_mask}
	\includegraphics[width=1.\textwidth,height=55pt]{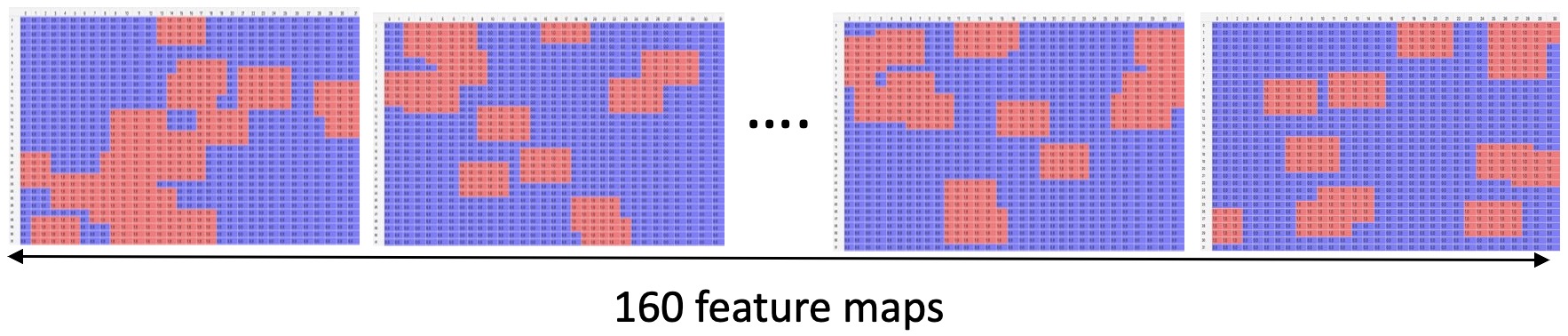}
	\caption{Mask sampling in \patchup{}.}
\end{subfigure}
\hfill
\begin{subfigure}{0.47\linewidth}
	\label{fig:cutmix_mask}
	\includegraphics[width=1.\textwidth]{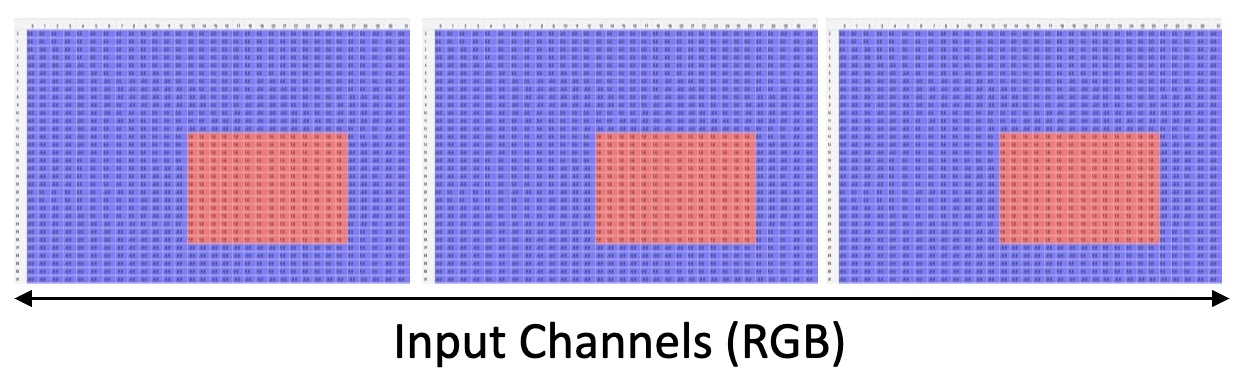}
	\caption{Mask sampling in \cutmix{}.}
\end{subfigure}
\caption{Mask sampling in \patchup{} is applied in the hidden state, compared to \cutmix{} which is applied in the input space. Red areas show the blocks that should be altered.}
\label{fig:mask_cmp}
\end{figure}

\begin{figure}[tbh]
\centering
\begin{minipage}{.51\textwidth}
  \centering
  \includegraphics[width=1.\linewidth, height=75pt]{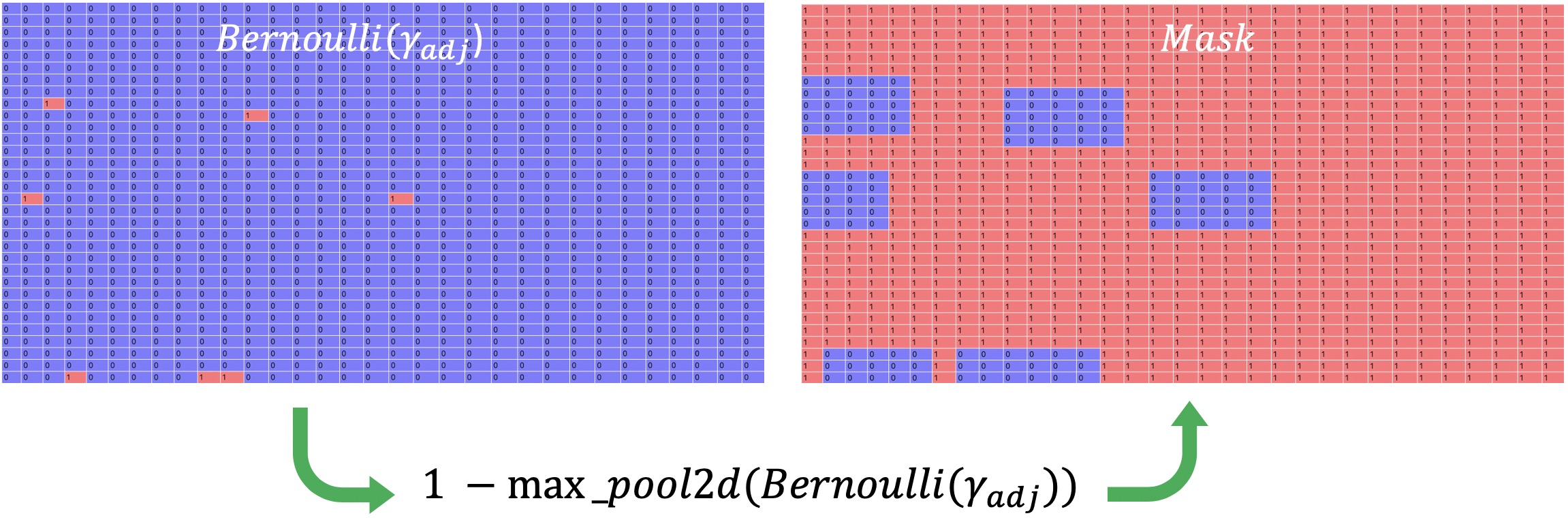} 
\caption {\patchup{} mask creation process ($block\_size = 5$). The left matrix shows the process of feature selection from feature maps. By using a $max\_pool2d$ function, we can create blocks around selected features. The $max\_pool2d$ function uses $stride = (1,\;1)$, $kernel\_size = (block\_size,\,block\_size)$, and $padding = (\frac{block\_size}{2}, \frac{block\_size}{2})$. Red and blue points are $1$ and $0$ in the generated binary mask, respectively.}
  \label{fig:block_selection}
\end{minipage}%
\hfill
\begin{minipage}{.47\textwidth}
  \centering
  \includegraphics[width=1.\textwidth]{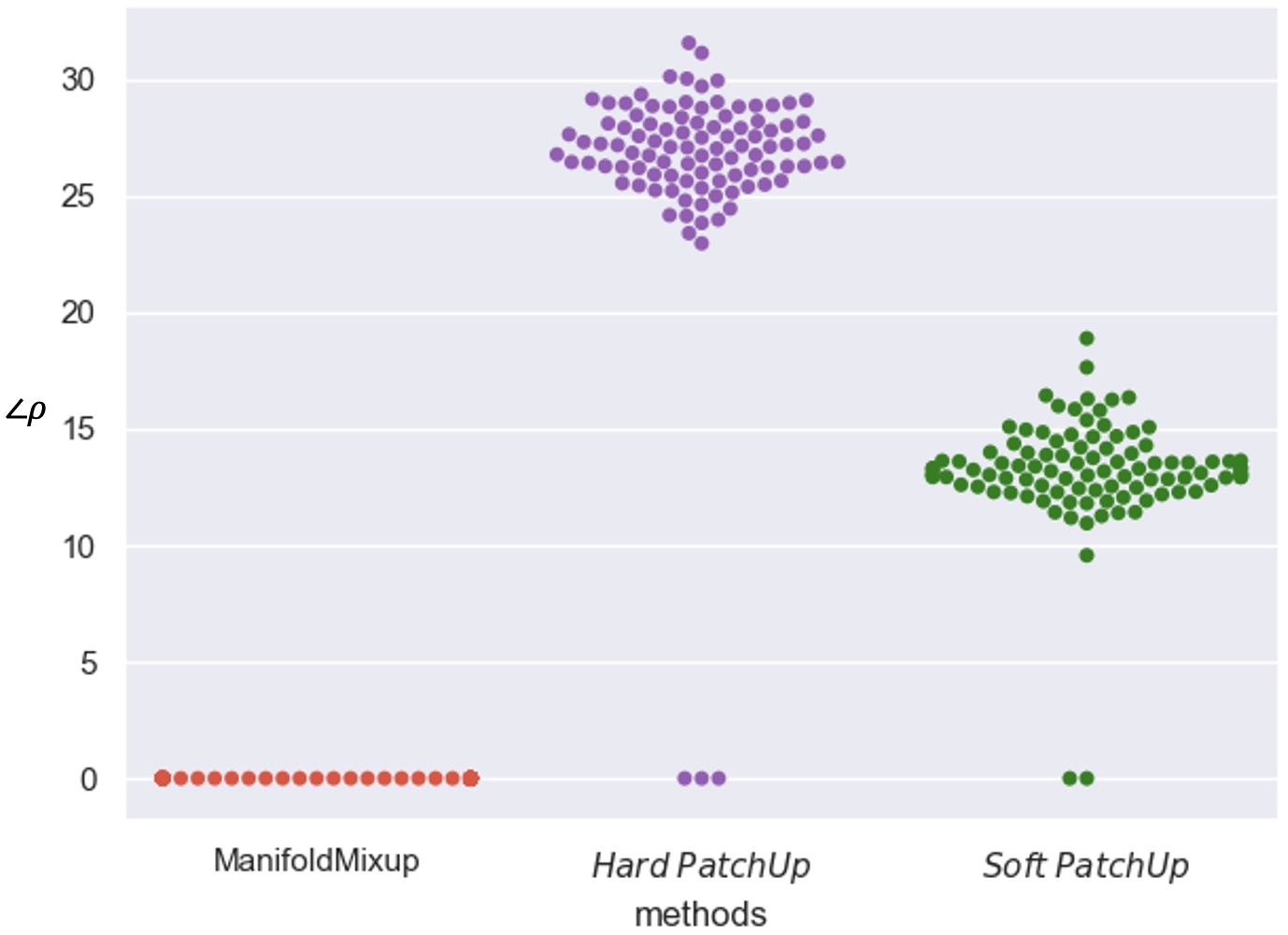} 
\caption{The comparison of $\rho$ for flattened hidden representations of a mini-batch of samples at the second residual block (layer $k = 3$) of WideResNet-28-10 with corresponding regularization method.}
\label{fig:sim_comparison}
\end{minipage}
\end{figure}

Figure~\ref{fig:block_selection} briefly illustrates and summarizes the binary mask creation process in \patchup{}. Lines 11 to 25 correspond to the interpolation and combination of hidden representations in the mini-batch in \patchup{}. Figure~\ref{fig:mask_cmp} compares the masks generated by \patchup{} and \cutmix{}. 

\section{Why random $k$?}
\label{appendix:random_k_selction}
\patchup{} applies block-level regularization at a randomly selected hidden representation layer $k$. The Information Bottleneck (IB) principle, introduced by \citet{DBLP:Tishby}, gives a formal intuition for selecting $k$ randomly. First, let us encapsulate the layers of the network into blocks where each block could contain more than one layer. Let $g_k$ be the $k$-th block of layers. In this case, sequential blocks share the information as a hidden representation to the next block of layers, sequentially. We can consider this case as a Markov chain of the block of layers as follows:
\begin{equation}
x \to g_{1}(x) \to  g_{2}(x) \to  g_{3}(x).
\end{equation}
In this scenario, the sequential communication between the intermediate hidden representations are considered to be an information bottleneck. Therefore, 
\begin{equation}
I(g_{3}(x);\;g_{2}(x)) < I(g_{2}(x);\;g_{1}(x)) < I(g_{1}(x);\;x),
\end{equation}
where $I(g_{k}(x);\;g_{k-1}(x))$ is the mutual information between the $k$-th and $(k-1)$-th layer.

If $g_{k=3}(x)$ has enough information to represent $x$, then applying regularization techniques in $g_{k=3}(x)$ will provide a better generalization to unseen data. However, most of the current state-of-the-art CNN models contain residual connections which break the Markov chain described above (since information can skip the $g_{k=3}$ layer). One solution to this challenge is to randomly select a residual block and apply regularization techniques like ManifoldMixup or \patchup{}. 

\section{\patchup{} Interpolation Policy Effect}
\label{appendix:sim_comparison}
\begin{figure}[H]
\centering
  \includegraphics[width=.32\textwidth]{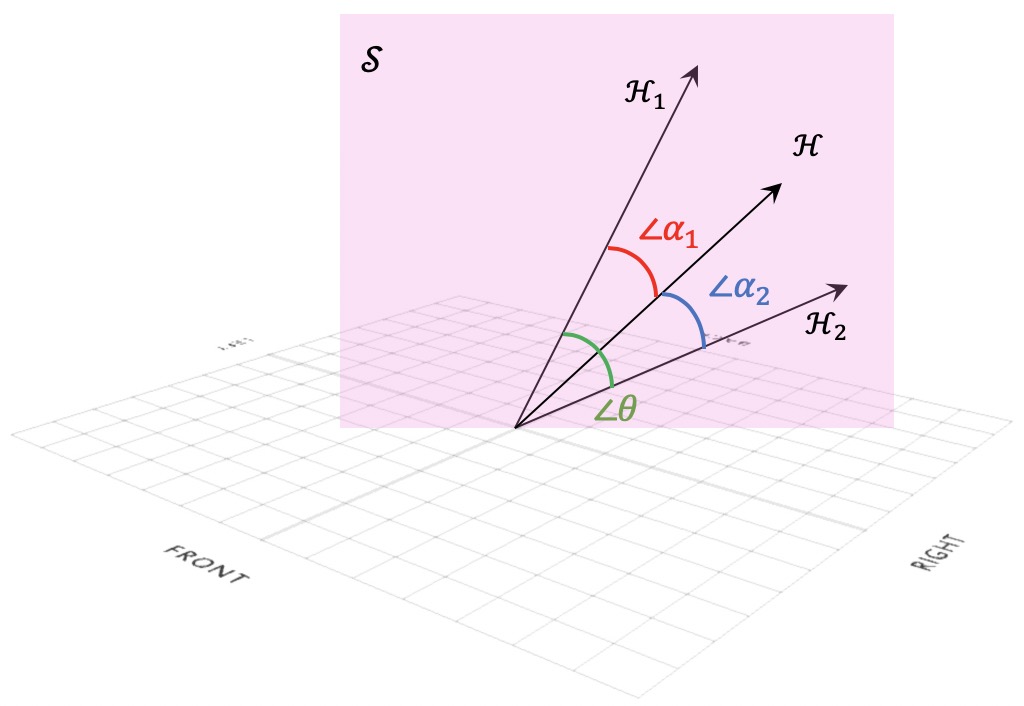} \hfill 
  \includegraphics[width=.32\textwidth]{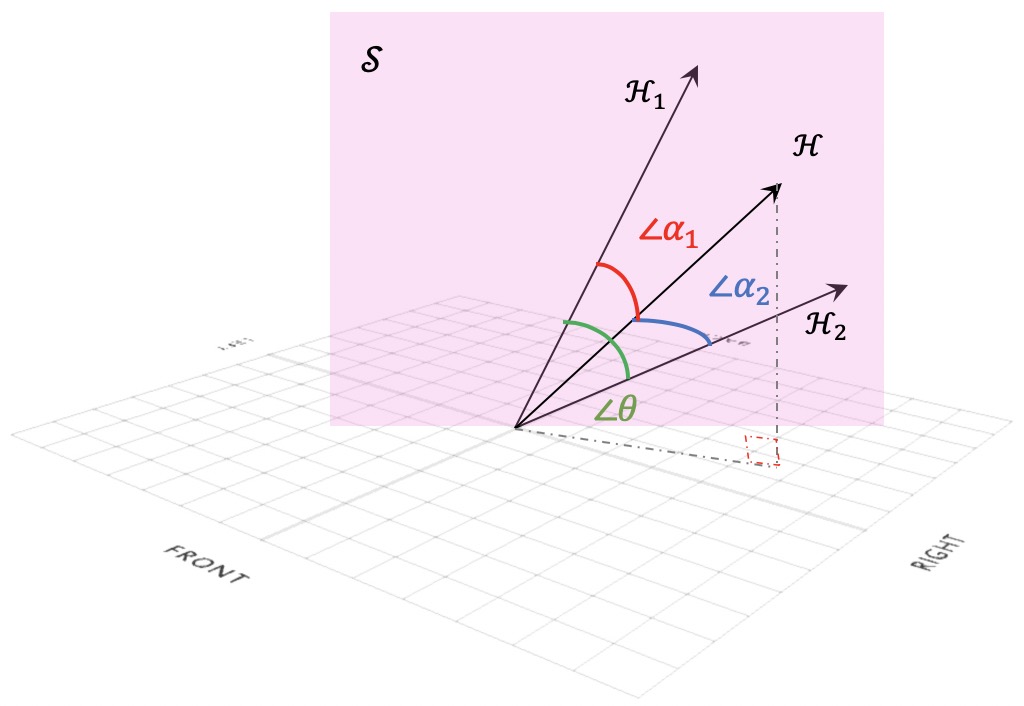} \hfill 
  \includegraphics[width=.32\textwidth]{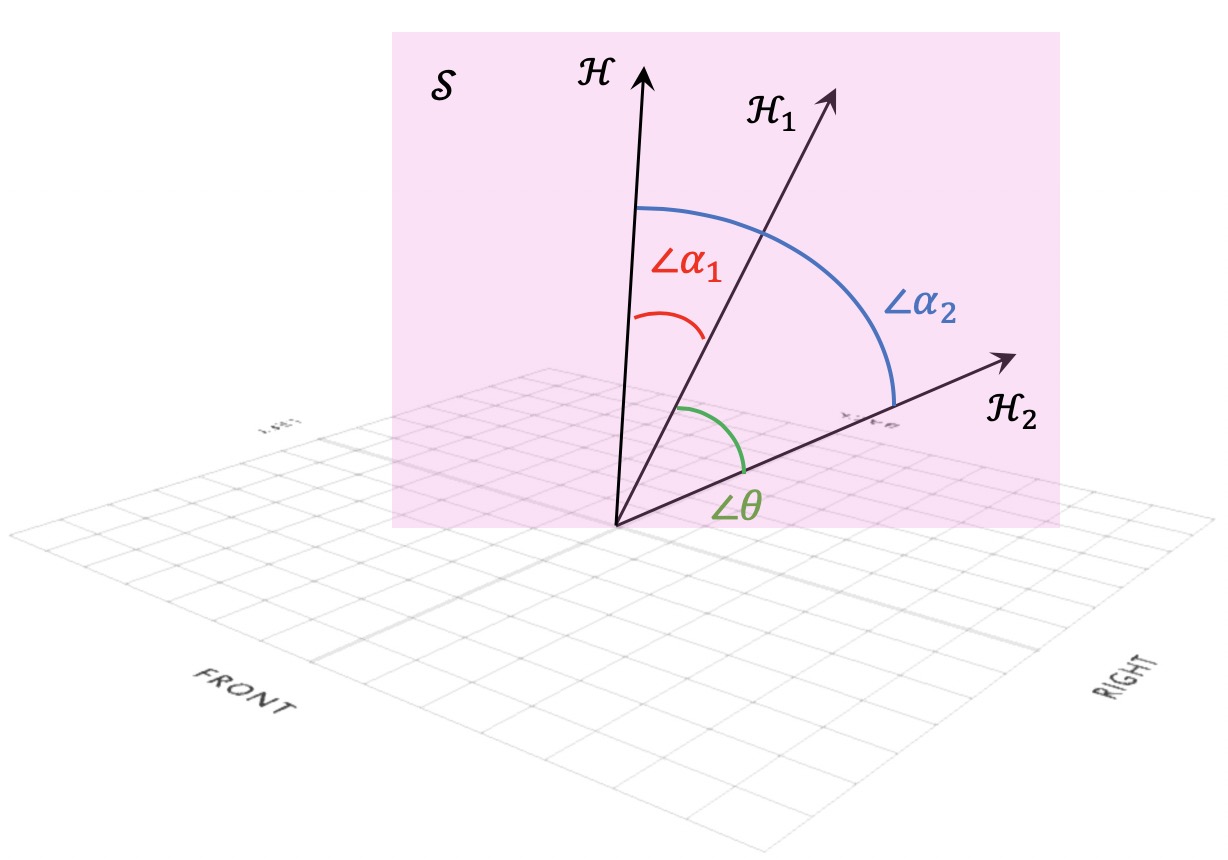}
    \caption{$\mathcal{H}_{1}$ and $\mathcal{H}_{2}$ are the flattened hidden representations. $\mathcal{H}$ is the flattened interpolated hidden representation that can be produced by either ManifoldMixup, \softpatchup{} or \hardpatchup{}.}
    \label{fig:sim_vectors}
\end{figure}

Assume that $\mathcal{H}_{1}$ and $\mathcal{H}_{2}$ are flattened hidden representations of two examples produced at layer $k$. And, $\mathcal{H}$ is the flattened interpolated hidden representation of these two paired samples at layer $k$. First, we calculate the cosine distance of the pairs ($\mathcal{H}_{2}$, $\mathcal{H}_{1}$), ($\mathcal{H}_{1}$, $\mathcal{H}$), and ($\mathcal{H}_{2}$, $\mathcal{H}$). Reversing the cosine of these cosine similarities give the angular distance between each pair of vectors denoted as $\angle\theta$, $\angle\alpha_{1}$, and $\angle\alpha_{2}$, respectively. There is always a surface that contains $\mathcal{H}_{2}$ and $\mathcal{H}_{1}$ denoted as $\mathcal{S}$. Mathematically, we have:
\begin{equation}
\begin{array}{l}
\label{eq:lsim_study}
    \angle \alpha_{1} = \cos^{-1}(\frac{\mathbf{\mathcal{H}_{1}} \cdot \mathbf{\mathcal{H}}}{\|\mathbf{\mathcal{H}_{1}}\|\|\mathbf{\mathcal{H}}\|}) \And
    \angle \alpha_{2} = \cos^{-1}(\frac{\mathbf{\mathcal{H}_{2}} \cdot \mathbf{\mathcal{H}}}{\|\mathbf{\mathcal{H}_{2}}\|\|\mathbf{\mathcal{H}}\|}) \And
    \angle \theta =
    \cos^{-1}(\frac{\mathbf{\mathcal{H}_{2}} \cdot \mathbf{\mathcal{H}_{1}}}{\|\mathbf{\mathcal{H}_{2}}\|\|\mathbf{\mathcal{H}_{1}}\|}),
\end{array}
\end{equation}
Let us define $\angle\rho=(\angle\alpha_{1}+ \angle\alpha_{2})-\angle\theta$ and $\angle\rho'= |\angle\alpha_{1}-\angle\alpha_{2}|-\angle\theta$.
According to the triangle inequality principle, either $\angle\rho$ or $\angle\rho'$ will be zero if, and only if, $\mathcal{H}\in\mathcal{S}$.
Figure~\ref{fig:sim_vectors} illustrates three possible scenarios for two paired flattened hidden representations and their flattened interpolated hidden representations. $\angle\rho$ and $\angle\rho'$ are zero for the left and right figures, respectively. We try to empirically show that $\mathcal{H}_{2}$, $\mathcal{H}_{1}$, and $\mathcal{H}$ always lie in the same surface $\mathcal{S}$ and $\mathcal{H}$ lies between $\mathcal{H}_{2}$ and $\mathcal{H}_{1}$ in ManifoldMixup. This means that $\angle\rho = 0$ for ManifoldMixup because of its linear interpolation policy. The middle figure in~\ref{fig:sim_vectors} is the case that both $\angle\rho$ and $\angle\rho'$ are not equal to zero. This figure shows that one possible situation is that flattened interpolated hidden representation does not lie in the surface $\mathcal{S}$. Our goal is to produce the interpolated hidden representation that lies in all possible places towards all dimensions in the hidden space. 
As discussed in section~\ref{para:patchup_vs_manifold}, ManifoldMixup can provide interpolated hidden representation only in a limited space. However, \softpatchup{} and \hardpatchup{} can produce a wide variety of interpolated hidden representations towards different dimensions. To support that, in WideResNet-28-10, for a mini-batch of 100 samples, we calculated the $\angle\rho$ for the flattened interpolated hidden representation produced by ManifoldMixup, \hardpatchup{}, and \softpatchup{} at the second residual block (layer $k=3$) with the same interpolation policy ($\lambda=.4$, $\gamma=.5$, and $block\_size=7$) for both \softpatchup{} and \hardpatchup{} for all samples in the mini-batch. The swarmplot~\ref{fig:sim_comparison} shows all $\angle\rho$ for the mini-batch are equal to zero in ManifoldMixup, which empirically supports our hypothesis. However, \patchup{} produces more diverse interpolated hidden representations towards all dimensions in the hidden space. It is worth mentioning that few $\angle\rho$ that are equal to zero in \softpatchup{} and \hardpatchup{} belong to the interpolated hidden representation that was constructed from the pairs with the same labels. 

\section{\patchup{} Experiment Setup and Hyper-parameter Tuning}
\label{appendix:hyperparameters}
We follow the ManifoldMixup in our experiment setup, we use SGD with 0.9 Nesterov momentum as an optimizer, mini-batch of 100, and weight decay of 1e-4~\cite{manifold}.

This section describes the hyper-parameters of each model in table~\ref{table:models_other_hp} following the hyper-parameter setup from ManifoldMixup \cite{manifold} experiments in order to create a fair comparison. First, we performed hyper-parameter tuning for the \patchup{} to achieve the best validation performance. Then we ran all the experiments five times, reporting the mean and standard deviation of errors and negative log-likelihoods for the selected models. We let models train for defined epochs and checkpoint the best model in terms of validation performance during the training. 
In our study, we used PreActResNet18, PreActResNet34, and WideResNet-28-10 models. Table~\ref{table:models_other_hp} shows the hyper-parameters used for training the models.

{\renewcommand{\arraystretch}{1.2}
\begin{table}[htbp!]
\centering
    \captionsetup[subtable]{position = top}
    \captionsetup[table]{position=top}
    \caption{Training time comparison between Puzzle Mix and \patchup{} using one GPU (V100) with 4 CPU cores and 32G of memory in our experiment setups. The reported time is rounded up to one hour.}
    \label{tab:tbl-running_time}
    \begin{subtable}{.85\linewidth}
        \resizebox{\linewidth}{!}{

\begin{tabular}{lcccccccc}
\toprule
                 & \multicolumn{2}{c}{CIFAR (hours)} &  & \multicolumn{2}{c}{SVHN (hours)} &  & \multicolumn{2}{c}{Tiny-ImageNet (hours)} \\ \cmidrule(r){2-3} \cmidrule(lr){5-6} \cmidrule(l){8-9} 
           & \shortstack[c]{\patchup{}}        & \shortstack[c]{Puzzle Mix}       &  & \shortstack[c]{\patchup{}}       & \shortstack[c]{Puzzle Mix}       &  & \shortstack[c]{\patchup{}}            & \shortstack[c]{Puzzle Mix}           \\ \cmidrule(r){2-3} \cmidrule(lr){5-6} \cmidrule(l){8-9}
\shortstack[l]{PreActResNet18}   & $18$             & $53$               &  & $18$            & $53$               &  & -                  & -                    \\
\shortstack[l]{PreActResNet34}   & $24$             & $64$               &  & $24$            & $60$               &  & -                  & -                    \\
\shortstack[l]{WideResNet-28-10} & $12$             & $26$               &  & $12$            & $32$               &  & $23$                 & $52$                   \\ \bottomrule
\end{tabular}}
\end{subtable}

\end{table}}
\begin{figure}[htbpt!]
\centering
\begin{subfigure}{0.45\linewidth}
	\includegraphics[width=1.\linewidth]{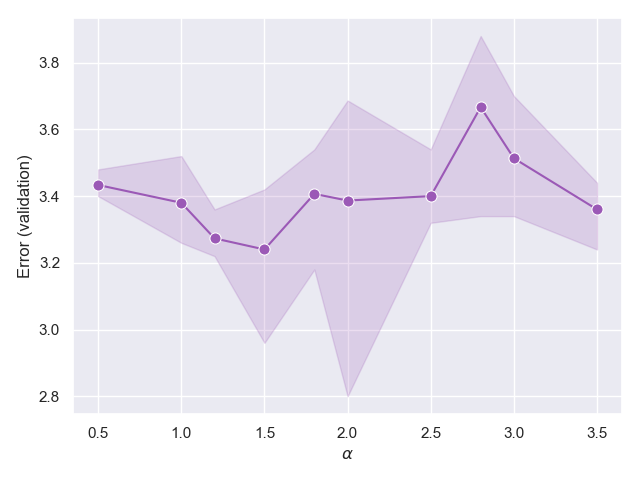}
	\caption{Impact of $\alpha$ in ManifoldMixup approach.}
	\label{fig:hyp_manifold}
	\end{subfigure}
\hfill
\begin{subfigure}{0.45\linewidth}
	\includegraphics[width=1.\linewidth]{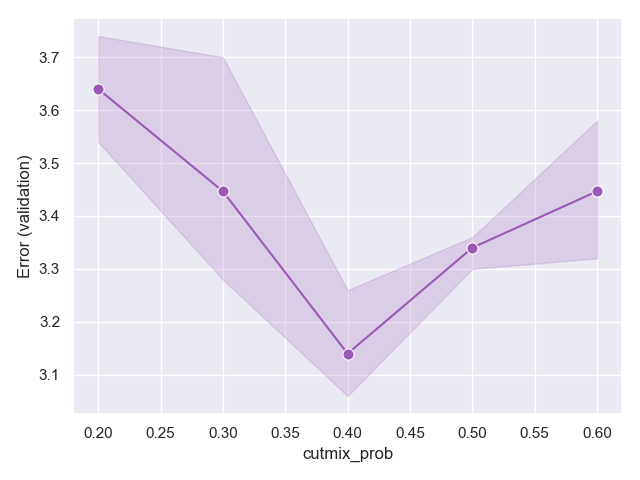}
	\caption{Impact of $cutmix\_prob$ in \cutmix{} approach.}
	\label{fig:hyp_cutmix}
	\end{subfigure}
\hfill
\begin{subfigure}{0.45\linewidth}
	\includegraphics[width=1.\linewidth]{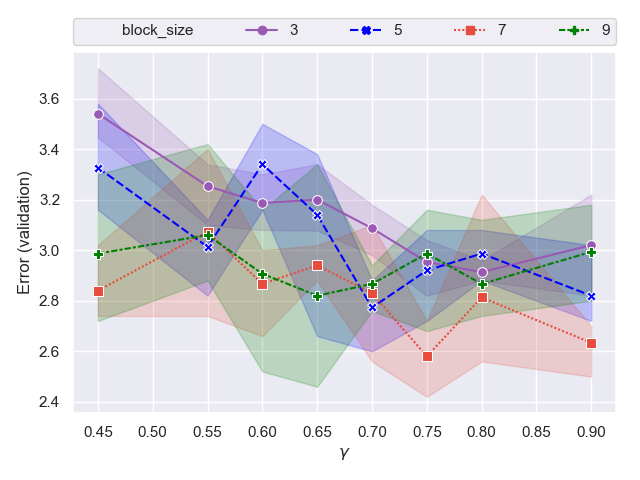} 
	\caption{Impact of $\gamma$, $block\_size$ with $patchup\_prob$ as 1.0 for \softpatchup{}.}
	\label{fig:hyp_soft}
	\end{subfigure}
\hfill
\begin{subfigure}{0.45\linewidth}
	\includegraphics[width=1.\linewidth]{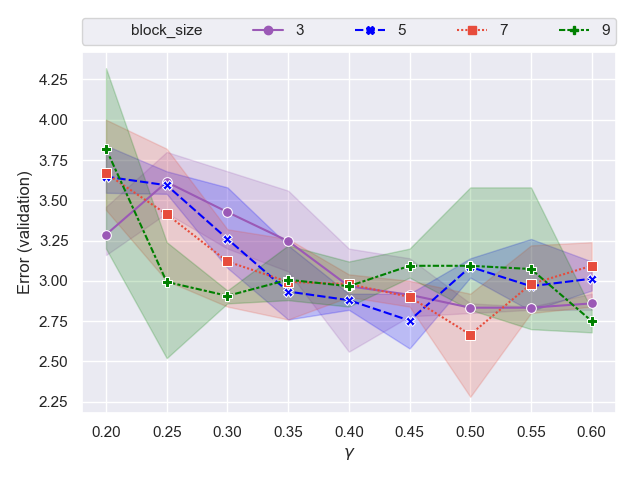} 
	\caption{Impact of $\gamma$, $block\_size$ with $patchup\_prob$ as 0.7 for \hardpatchup{}.}
	\label{fig:hyp_hard}
	\end{subfigure}
\caption{Impact of hyper-parameters $\gamma$, $block\_size$ and $patchup\_prob$ on error rates in the CIFAR10 validation set for PreActResNet18. We repeated each job three times to collect the mean and the standard deviation of errors. Marked points are the mean of the error rate in the validation set. And, the shadow shows the bootstrapping of results for each hyper-parameter setting.
The lower numbers on the y-axes correspond to better performance.}
\label{hyptune}
\end{figure}
{\renewcommand{\arraystretch}{1.}
\begin{table*}[t]
\centering
    \begin{subtable}{.6\linewidth}
        \resizebox{\linewidth}{!}{
	\begin{tabular}[h]{ ll rrr }
		\toprule
		{Model} & {\textit{lr}} & {\textit{lr} steps} & {step factor} & {Epochs}\\
        \midrule
		{PreactResnet18}  & $0.1$ & $500${-}$1000${-}$1500$ & $0.1$ & $2000$\\
		\cmidrule(lr){2-5}
		{PreactResnet34} & $0.1$ &$500${-}$1000${-}$1500$ & $0.1$ & $2000$\\
		\cmidrule(lr){2-5}
		{WideResnet-28-10} & $0.1$ & $200${-}$300$ & $0.1$ & $400$\\
        \bottomrule
	\end{tabular}}
\end{subtable}
 \captionsetup[subtable]{position = bottom}
    \captionsetup[table]{position=bottom}
    \caption{The hyper-parameters used for each model to compare the effect of each regularization technique. The learning rate is denoted as \textit{lr}. And, \textit{lr} is multiplied at each learning rate schedule step by step factor.}
    \label{table:models_other_hp}
\end{table*}}

\patchup{} adds $patchup\_prob$, $\gamma$ and $block\_size$ as hyper-parameters. $patchup\_prob$ is the probability that the \patchup{} operation is performed for a given mini-batch, i.e if there are $N$ mini-batches and $patchup\_prob$ is $p$, \patchup{} is performed in $p$ fraction of $N$ mini-batches. $\gamma$ and $block\_size$ are described in section~\ref{patchup}. 
We tuned the \patchup{} hyper-parameter on CIFAR10 with the PreActResNet18. To create a validation set, we split $10\%$ of training samples into a validation set. We set $\alpha$ to $2$ in \patchup{}. For \softpatchup{}, we set $patchup\_prob$ to $1.0$ and applied \patchup{} to all mini-batches in training. Then, we did a grid search by varying $\gamma$ from $0.45$ to $0.9$ and $block\_size$ from $3$ to $9$. We found that $\gamma$ of $0.75$ and $block\_size$ of $7$ work best for \softpatchup{} as shown in figure~\ref{fig:hyp_soft}. Similarly, for \hardpatchup{}, we set $patchup\_prob$ to $0.7$ and performed a grid search by varying $\gamma$ from $0.2$ to $0.6$ and $block\_size$ from $3$ to $9$. We found that $block\_size$ of $7$ and $\gamma$ of $0.5$ yield the best results for \hardpatchup{} as shown in figure~\ref{fig:hyp_hard}. Figure~\ref{fig:hyp_manifold} shows that ManifoldMixup with ($\alpha = 1.5$) achieves the best validation performance. For cutout, we used the same hyper-parameters proposed in \cite{cutout}, setting cutout to $16$ for CIFAR10, $8$ for CIFAR100, and $20$ for SVHN following~\cite{cutout}.
Figure~\ref{fig:hyp_cutmix} shows that \cutmix{} achieves its best validation performance in PreActResNet18 in CIFAR10 with $cutmix\_prob = 0.4$. Furthermore, DropBlock achieves its best validation performance on this task by setting the block size and $\gamma$ to $7$ and $0.9$, respectively~\cite{dropblock}. 
To train models using regularization methods, we use one GPU (v100). 
The training time for other methods is almost the same as \patchup{} since they are not adding any computational overhead. But, Puzzle Mix adds about three times and two times more computational overhead in SVHN, CIFAR, and Tiny-ImageNet, respectively. Table~\ref{tab:tbl-running_time} show the training time comparison. 

As we showed in our ablation study (shown in Figure~\ref{fig:layer_zoro} and Table~\ref{tab:layer_study_k}). Applying the PatchUp Mask in input space drastically destroys the semantic concepts in the input images (Figure\ref{fig:layer_zoro}). Also worth mentioning is that it is unlikely that applying the PatchUp Mask in latent space leads to drastically changing the hidden representation since we have many channels. Tuning gamma as a hyperparameter will prevent such large semantic changes in the latent space. However, gamma cannot avoid such negative effects in input space. As per Table~\ref{tab:layer_study_k}, applying the CutMix Binary Mask in input space but still using the PatchUp learning objective leads to a performance improvement of $0.836$ when training WRN-28-10 on CIFAR-100, which is significant when considering the tight competition between regularization methods.

\section{
Further explanation on \patchup{} and its learning objective:}
\label{appendix:learning_objective_details}
According to the Eq.~\ref{equ:mix_lam} ${Mix}_{p_{u}}$ is computed as follows:

$$Mix_{p_{u}}[a, b]=p_{u} a+\left(1-p_{u}\right) b$$

In the first term in equation~\ref{eq:swap_loss}, we take a convex combination of the following two loss functions: one with respect to the unchanged part of the feature maps ($(p_{u})$) and the other with respect to the interpolated part ($(1 - {p_{u}})$). Please note that the loss function for the interpolated part will change based on whether we do Soft or Hard PatchUp. Therefore, we can expand the PatchUp learning objective as follows:

\begin{align*}
L(f)=& E_{\left(x_{i}, y_{i}\right) \sim P} E_{\left(x_{j}, y_{j}\right) \sim P} E_{\lambda \sim \operatorname{Beta}(\alpha, \alpha)} E_{k \sim S} M i x_{p_{u}}\left[\ell\left(f_{k}\left(\phi_{k}\right), y_{i}\right), \ell\left(f_{k}\left(\phi_{k}\right), Y\right)\right] \\
&~~~~~~~~~~~~~~~~~~~~~~~~~~~~~~~~~~~~~~~~~~~~~~~~~~~~~~~~~~~ +\ell\left(f_{k}\left(\phi_{k}\right), W\left(y_{i}, y_{j}\right)\right)
\end{align*}

Based on the operation defined in section 2, we can substitute and expand the learning objective for Hard PatchUp as follows: (For Soft PatchUp the process will be similar.) In the Hard PatchUp:
$$
Y = y_j
$$
And according to Equation~\ref{eq:target_reweighted},
$$
W_{{hard }}(y_{i}, y_{j})={Mix}_{p_{u}}(y_{i}, y_{j}) = p_{u} y_i + (1 -p_{u})y_j
$$

$$
{Mix}_{p_{u}}[\ell(f_{k}(\phi_k),y_i),\ell(f_{k}(\phi_k),Y)]={p_{u}} \ell(f_{k}(\phi_k),y_i) + (1 - {p_{u}}) \ell(f_{k}(\phi_k),y_j)
$$

\noindent Therefore, we can conclude the learning objective for Hard PatchUp is defined as follows:

$$
L(f) = {p_{u}} \ell(f_{k}(\phi_k),y_i) + (1 - {p_{u}}) \ell(f_{k}(\phi_k),y_j) + \ell\left(f_{k}\left(\phi_{k}\right), p_{u} y_i + (1 -p_{u})y_j\right)
$$
where the $
\phi_{ {hard }}(g_{k}(x_{i}), g_{k}(x_{j}))={M} \odot g_{k}(x_{i})+({1}-{M}) \odot g_{k}(x_{j})
$.
\hardpatchup{} is a fully differentiable operation. The masks for the hard PatchUp are sampled outside the computation graph and passed as input to the architecture. Hence the entire architecture is differentiable end-to-end. This is similar to the reparameterization trick that was proposed in~\cite{kingma2013auto}. A similar trick was also applied in Variational Dropout~\cite{kingma2015variational} and other similar methods as well.
Similarly, we can substitute the operations (${Mix}$ and $\phi_{ {soft }}$ defined in Equation 3 and 5) to expand the learning objective of Soft PatchUp.

PatchUp works by applying a structured noisy process to the representations at a random point in the forward pass (random layer). The model then learns to recognize the structure in the noisy representations created in the forward pass by back-propagating through the model with the gradients from the PatchUp learning objective.  Since all operations are differentiable and involve either swapping or mixing latent spaces (similar to the reparameterization trick), we will not have any non-differentiability issues when calculating the gradients. Hence, deriving the backprop manually is not necessary.

\section{ Generalization on Deformed Images}
\label{appendix:affine_transformation}
{\renewcommand{\arraystretch}{1.1}
\begin{table*}[htbp!]
\centering
    \captionsetup[subtable]{position = bottom}
    \captionsetup[table]{position=bottom}
    \caption{Error rates in the test set on samples subject to affine transformations for WideResNet-28-10 trained on CIFAR-100 with indicated regularization method. We repeated each test for five trained models to report the mean and the standard deviation of errors. The best performance result is shown in bold, second best is underlined. The lower number is better.}
    \label{tab:affine_result}
    \begin{subtable}{1.0\linewidth}
        \resizebox{\linewidth}{!}{
    \begin{tabular}{lllllll}
    \toprule
    \shortstack[l]{Transformation}      & \shortstack[l]{cutout}           & \shortstack[l]{\cutmix{}}          & \shortstack[l]{ManifoldMixup}  & \shortstack[l]{Puzzle Mix}  & \shortstack[l]{\softpatchup{}}              & \shortstack[l]{\hardpatchup{}}   \\ \hline
    \shortstack[l]{Rotate (-20, 20)}    & $36.162 \pm 0.633 $ & $34.236 \pm 0.785 $ & $35.774 \pm 0.621 $ & $28.35 \pm 0.561$  & $\mathbf{31.282 \pm 0.622} $ & \ul{$31.340 \pm 0.318$} \\
    \shortstack[l]{Rotate (-40, 40)}    & $57.220 \pm 0.549 $ & $56.512 \pm 0.752 $ & $56.610 \pm 0.877 $ & $51.49 \pm 1.028$  & $\mathbf{52.014 \pm 0.916} $ & \ul{$53.804 \pm 0.576$} \\
    \shortstack[l]{Shear (-28.6, 28.6)} & $33.482 \pm 0.463 $ & $31.770 \pm 0.312 $ & $32.300 \pm 0.317 $ &  $27.178 \pm 0.256$ & \ul{$30.898 \pm 0.836$} & $\mathbf{28.426 \pm 0.430} $ \\
    \shortstack[l]{Shear (-57.3, 57.3)} & $53.328 \pm 0.587 $ & \ul{$50.618 \pm 0.552$} & $52.366 \pm 0.170 $ &  $47.414 \pm 0.482$ & $51.908 \pm 0.632 $ & $\mathbf{48.334 \pm 0.631} $ \\
    \shortstack[l]{Scale (0.6)}         & $56.770 \pm 0.376 $ & $\mathbf{45.980 \pm 0.404} $ & $63.924 \pm 2.160 $ & $46.354 \pm 0.869$ &  $52.648 \pm 0.616 $ & \ul{$46.924 \pm 1.035$} \\
    \shortstack[l]{Scale (0.8)}         & $30.550 \pm 0.611 $ & \ul{$26.818 \pm 0.328$} & $29.012 \pm 0.372 $ & $24.428 \pm 0.171$  & $27.188 \pm 0.597 $ & $\mathbf{23.840 \pm 0.535} $ \\
    \shortstack[l]{Scale (1.2)}         & $47.268 \pm 0.639 $ & $51.258 \pm 0.817 $ & $\mathbf{41.644 \pm 0.846} $ &$49.18 \pm 1.046$  &  \ul{$42.108 \pm 0.985$} & $43.370 \pm 1.223 $ \\
    \shortstack[l]{Scale (1.4)}         & $79.000 \pm 0.933 $ & $82.562 \pm 0.575 $ & \ul{$72.752 \pm 0.846$} & $82.152 \pm 0.41$  & $\mathbf{70.970 \pm 1.433} $ & $77.370 \pm 1.457 $ \\ \bottomrule
    \end{tabular}}
\end{subtable}

\end{table*}
}

We created the deformed test sets from CIFAR100, as described in Section~\ref{exp:deformed_images_test}. Table~\ref{tab:affine_result} shows improved quality of representations learned by a WideResNet-28-10 model regularized by \patchup{} on CIFAR100 deformed test sets. The significant improvements in generalization provided by \patchup{} in this experiment shows the high quality of representations learned with \patchup{}.
\section{Robustness to Adversarial Examples}
\label{appendix:adversarial-attack}
The adversarial attacks refer to small and unrecognizable perturbations on the input images that can mislead deep learning models \cite{goodfellow2014FGSM, goodfellow2016deep}. 
One approach to creating adversarial examples is using the Fast Gradient Sign Method (FGSM), also known as a white-box attack~\cite{goodfellow2014FGSM}. FGSM creates examples by adding small perturbations to the original examples. Once a regularized model is trained then FSGM creates adversarial examples as follows \cite{goodfellow2014FGSM}:
\begin{equation}
\begin{array}{l}
\label{eq:fgsm_perturbation}
    x^{\prime}= x + \epsilon \times \operatorname{sign}\left(\nabla_{\boldsymbol{x}} J(\boldsymbol{\theta}, \boldsymbol{x}, y)\right).
\end{array}
\end{equation}
where $x^{\prime}$ is an adversarial example, $x$ is the original example, $y$ is the ground truth label for $x$, and $J(\boldsymbol{\theta}, \boldsymbol{x}, y)$ is the loss of the model with parameters of $\theta$. $\epsilon$ controls the perturbation. 

\begin{figure}[bt!]
\centering
\begin{subfigure}{0.46\linewidth}
\centering
	\includegraphics[width=.85\linewidth]{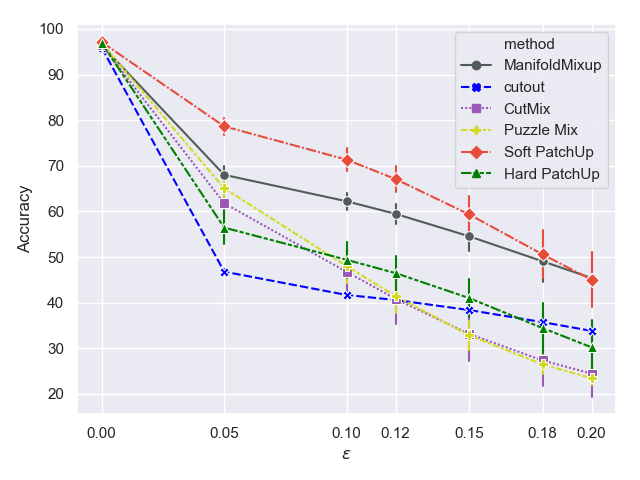} 
	\caption{PreActResNet18 in CIFAR10.}
	\label{fig:appendix_attack_c10_p18}
	\end{subfigure}
\hfill
\begin{subfigure}{0.46\linewidth}
\centering
	\includegraphics[width=.85\linewidth]{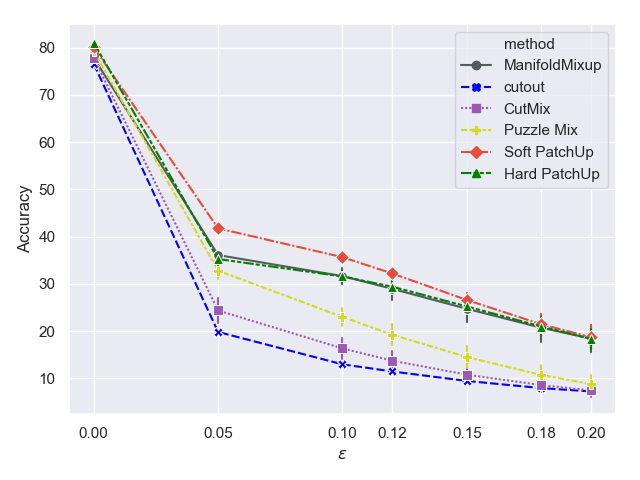} 
	\caption{PreActResNet18 in CIFAR100.}
	\label{fig:appendix_attack_c100_p18}
	\end{subfigure}
\hfill
\begin{subfigure}{0.46\linewidth}
\centering
	\includegraphics[width=.85\linewidth]{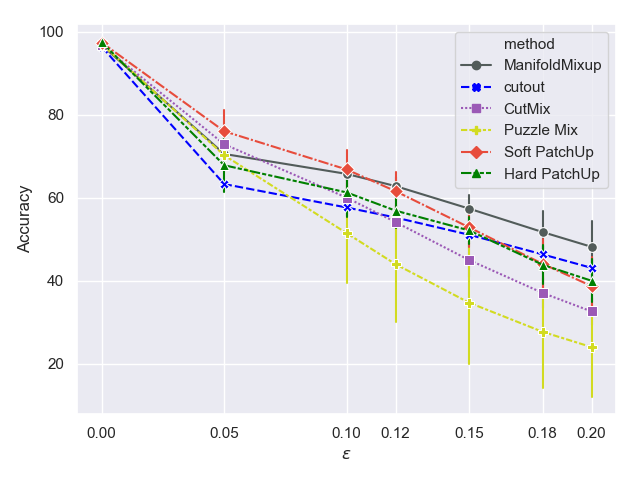} 
	\caption{PreActResNet34 in CIFAR10.}
	\label{fig:appendix_attack_c10_p34}
	\end{subfigure}
\hfill
\begin{subfigure}{0.46\linewidth}
    \centering
	\includegraphics[width=.85\linewidth]{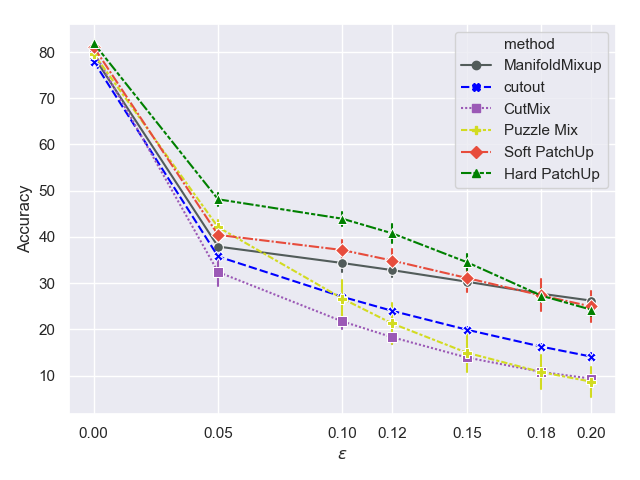} 
	\caption{PreActResNet34 in CIFAR100.}
	\label{fig:appendix_attack_c100_p34}
	\end{subfigure}
\hfill
\begin{subfigure}{0.46\linewidth}
\centering
	\includegraphics[width=.85\linewidth]{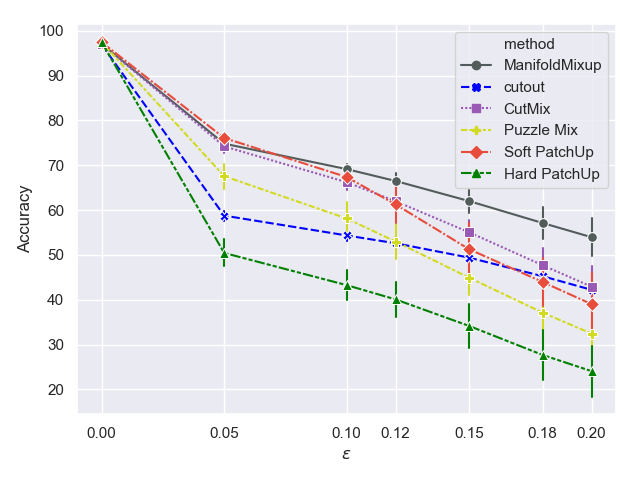} 
	\caption{WideResNet28-10 in CIFAR10.}
	\label{fig:appendix_attack_c10_wrn}
	\end{subfigure}
\hfill
\begin{subfigure}{0.46\linewidth}
\centering  
	\includegraphics[width=.85\linewidth]{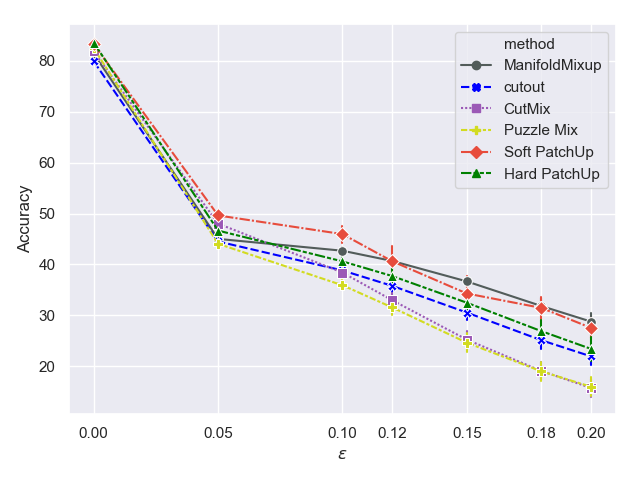}
	\caption{WideResNet28-10 in CIFAR100.}
	\label{fig:appendix_attack_c100_wrn}
	\end{subfigure}
\hfill
\begin{subfigure}{0.46\linewidth}
\centering
	\includegraphics[width=.85\linewidth]{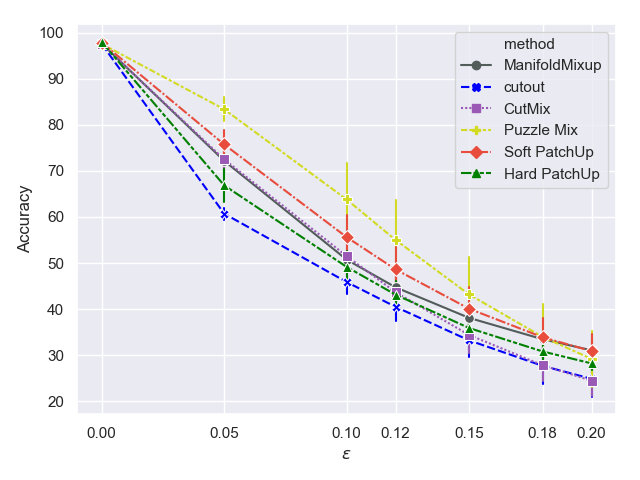}
	\caption{WideResNet28-10 in SVHN.}
	\label{fig:appendix_attack_svhn_wrn}
\end{subfigure}	
\hfill
\begin{subfigure}{0.46\linewidth}
\centering
	\includegraphics[width=.85\linewidth]{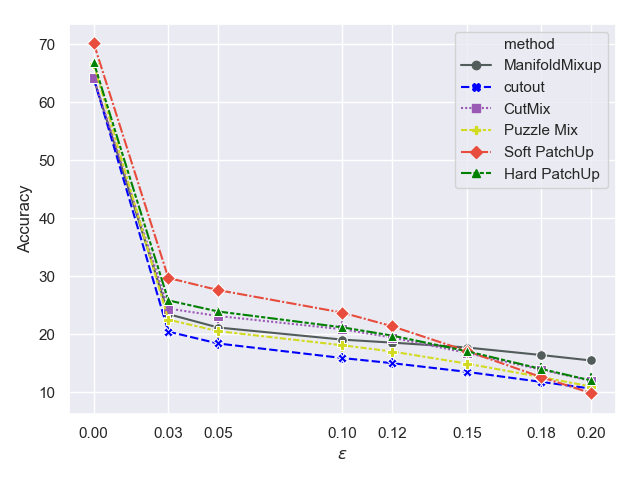}
	\caption{WideResNet28-10 in Tiny-ImageNet.}
	\label{fig:appendix_attack_tiny_wrn}
\end{subfigure}	
\caption{Robustness to the FGSM attack, known as a white-box attack. We repeated each test for five trained models to report the mean and the standard deviation of the accuracy of each method against the FGSM attack. The higher values on the y-axes show the robustness of the model against the attack. And, $\epsilon$ is the magnitude that controls the perturbation. }
\label{fig:appendix_adv_attack}
\end{figure}

Our experiments show the effectiveness of \softpatchup{} against the attacks in most cases. However, \hardpatchup{} performed well against the FGSM attack only on PreActResNet34 for CIFAR100. Figure~\ref{fig:appendix_adv_attack} shows the comparison of the state-of-the-art regularization techniques' effect on model robustness against the FGSM attack.

\section{Analysis of \patchup{}'s Effect on Activations}
\label{appendix:activation_comparison}
\begin{figure}[htbp!]
\centering
\begin{subfigure}{0.47\textwidth}
	\includegraphics[width=1.\textwidth]{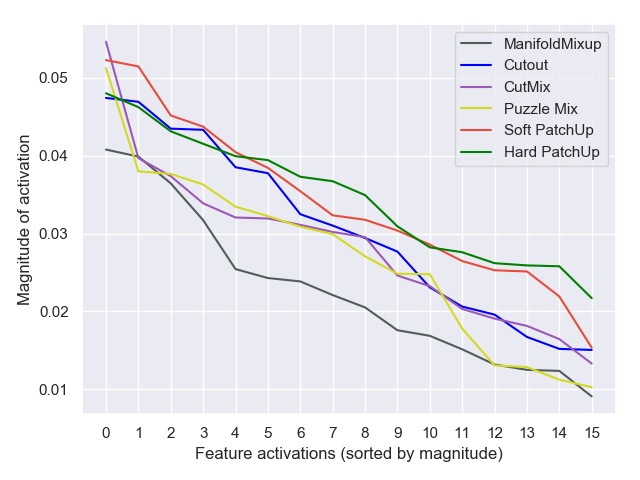}
	\caption{Comparison on the first convolution module.}
	\label{fig:appendix_activation_conv1}
\end{subfigure}
\hfill
\begin{subfigure}{0.47\textwidth}
	\includegraphics[width=1.\textwidth]{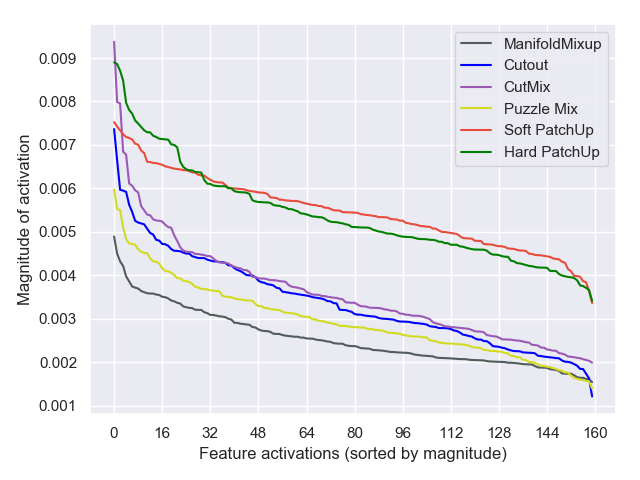} \caption{Comparison on 1st Residual Block.}
	\label{fig:appendix_activation_layer1}
\end{subfigure}
\hfill
\begin{subfigure}{0.47\textwidth}
	\includegraphics[width=1.\textwidth]{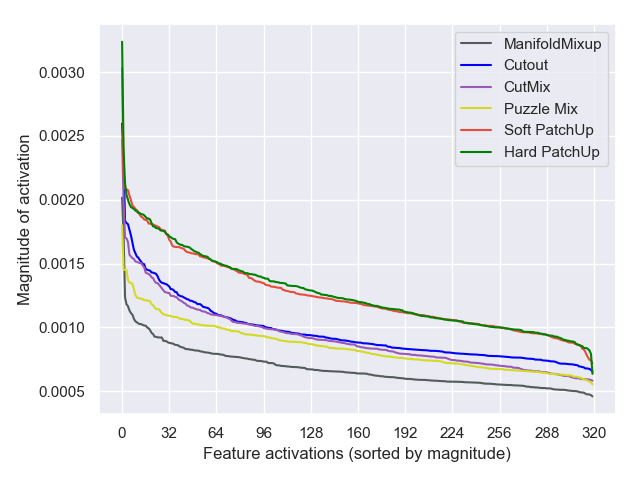} \caption{Comparison on 2nd Residual Block.}
	\label{fig:appendix_activation_layer2}
\end{subfigure}
\hfill
\begin{subfigure}{0.47\textwidth}
	\includegraphics[width=1.\textwidth]{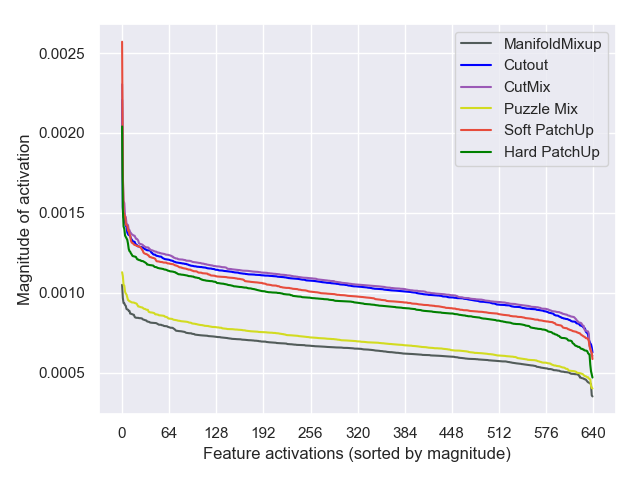}
	\caption{Comparison on 3rd Residual Block.}
	\label{fig:appendix_activation_layer3}
\end{subfigure}
\hfill
\caption{The effect of the state-of-the-art regularization techniques on activations in WideResNet28-10 for CIFAR100 test set. Each curve is the magnitude of feature activations, sorted by descending value, and averaged over all test samples for each method. The higher magnitude shows a wider variety of the produced features by the model at each block.}
\label{fig:appendix_activation_cmp}
\end{figure}

In our implementation WideResNet28-10 has a conv2d module followed by three residual blocks. Figure~\ref{fig:appendix_activation_cmp} illustrates the comparison of ManifoldMixup, cutout, CutMix, \softpatchup{}, and \hardpatchup{}. Figure~\ref{fig:appendix_activation_conv1},~\ref{fig:appendix_activation_layer1}, and~\ref{fig:appendix_activation_layer2} show that \patchup{} produces more variety of features in layers that we apply \patchup{} on. 

\begin{figure}[htbp!]
\centering
\begin{subfigure}{0.49\textwidth}
	\includegraphics[width=1.\textwidth]{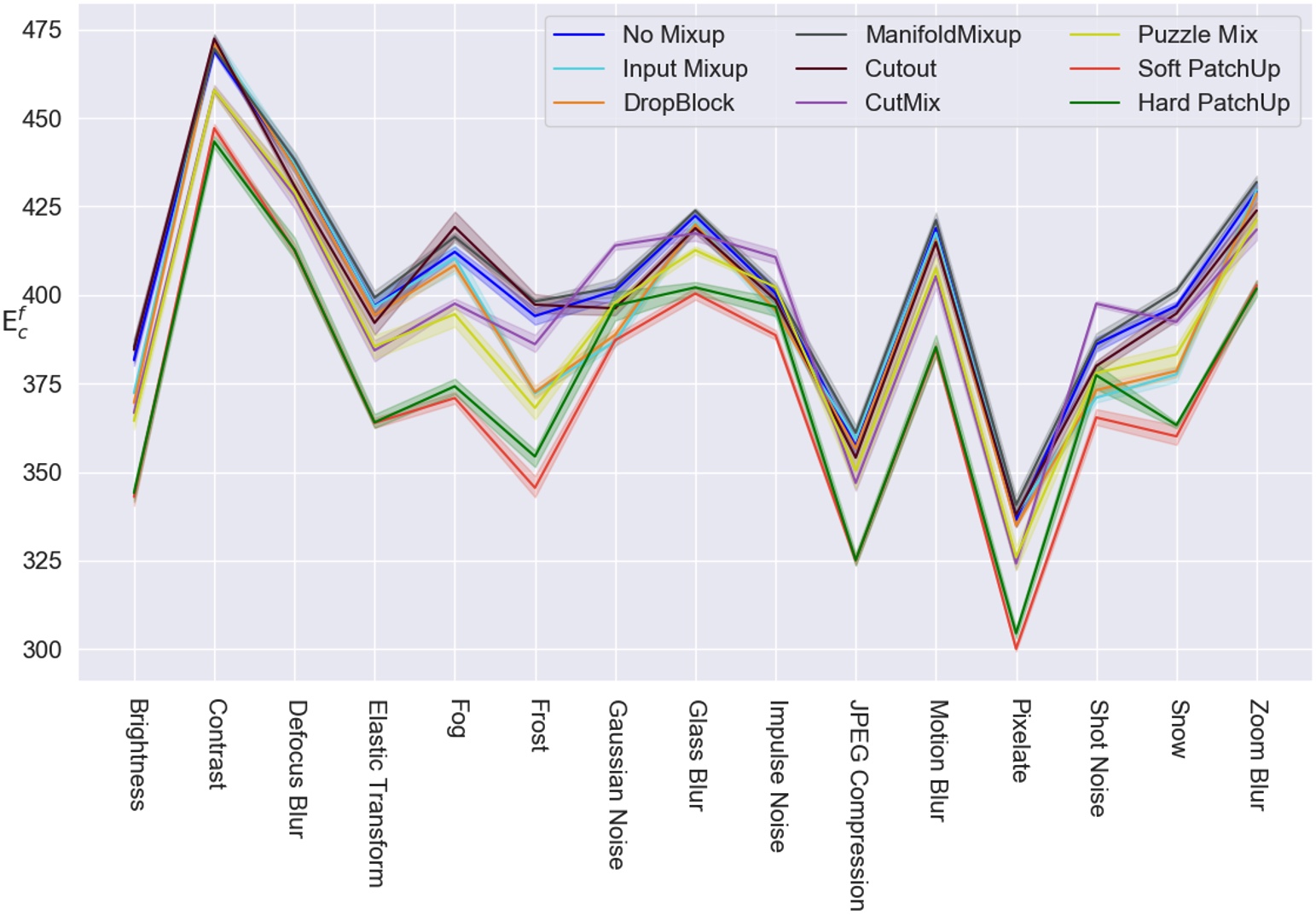}
	\caption{Comparison on ResNet101.}
	\label{fig:appendix_tiny_c_101}
\end{subfigure}
\hfill
\begin{subfigure}{0.49\textwidth}
	\includegraphics[width=1.\textwidth]{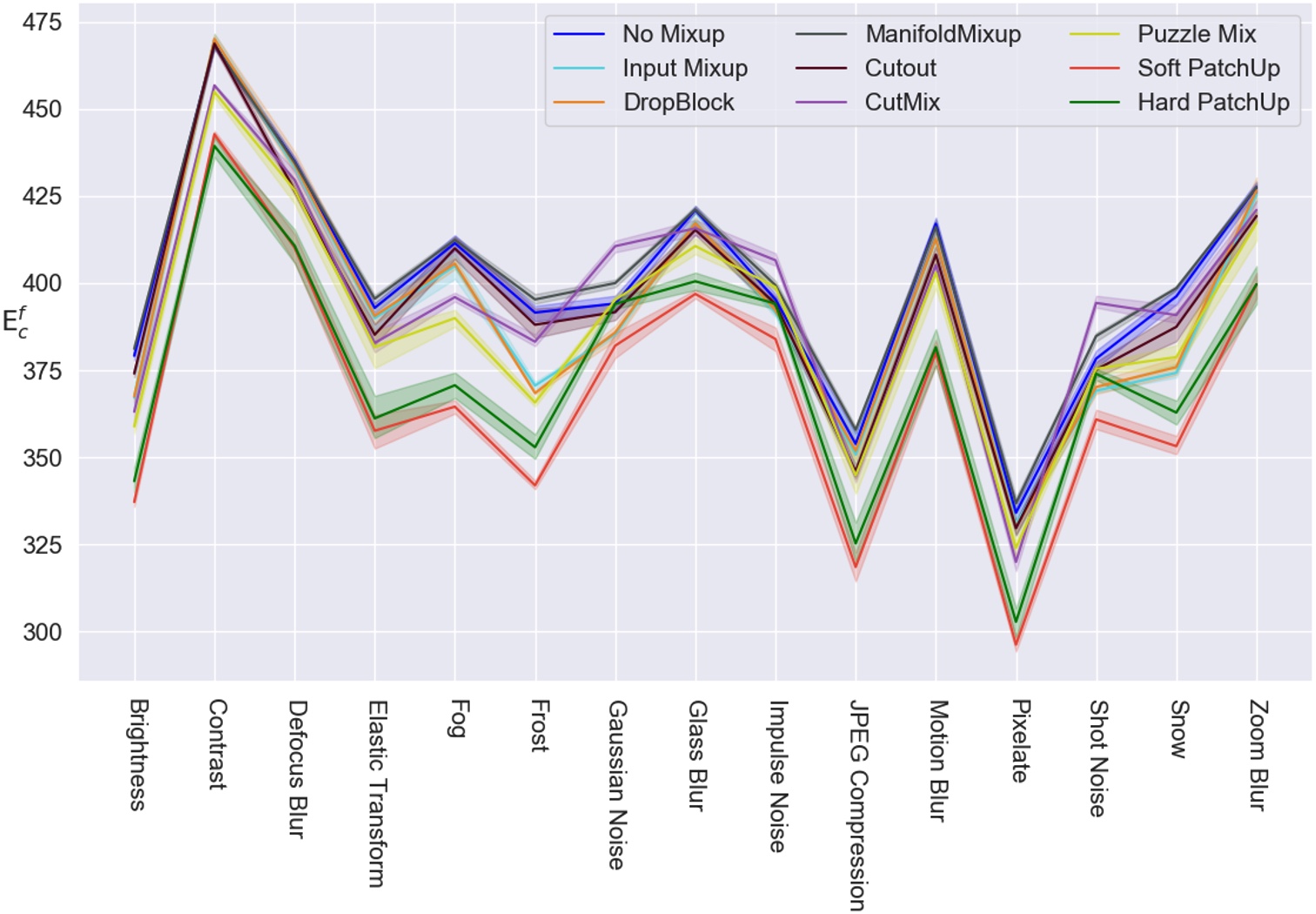} \caption{Comparison on on ResNet152.}
	\label{fig:appendix_tiny_c_152}
\end{subfigure}
\hfill
\begin{subfigure}{0.49\textwidth}
	\includegraphics[width=1.\textwidth]{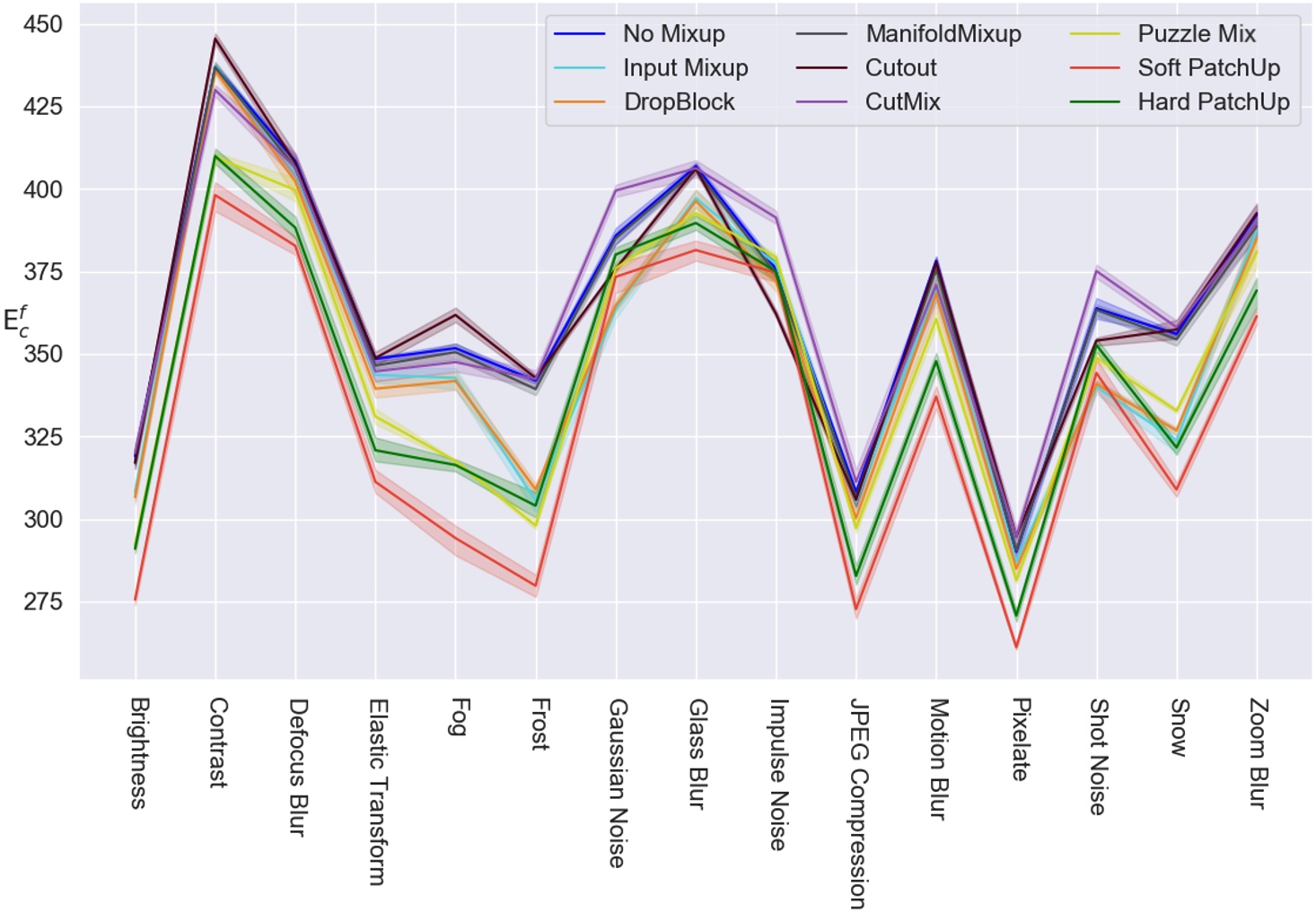} \caption{Comparison on on WideResNet-28-10.}
	\label{fig:appendix_tiny_c_wrn}
\end{subfigure}
\caption{Robustness to Tiny-ImageNet Common Corruptions (Tiny-ImageNet-C). We repeated each test for five trained models to report the mean and the standard. The lower values on the y-axes show the robustness of the model against the input common corruptions. The y-axis is the sum of error rates for each category. And, the x-axis represents the list of corruptions in Tiny-ImageNet-C.}
\label{fig:appendix_tiny_imagenet_c}
\end{figure}

\section{
More Details on Image Classification results}
\label{appendix:calssification_details}
This section contains the image classification error and Test NLL on CIFAR10, CIFAR100, and SVHN datasets. Tables~\ref{tab:tbl-cifar} and~\ref{tab:tbl-svhn} shows Test error rates and test NLL on CIFAR10, CIFAR100, and SVHN datasets using PreActResNet18, PreActResNet34, and WideResNet-28-10 models. 

When we have a data-scarce scenario, high-capacity models tend to memorize the examples. This leads to models with reasonable testing performance but that are very sensitive to small changes in test distribution as shown in Tiny-ImageNet-C experiments. Tiny-ImageNet has 200 classes while CIFAR-100 has 100 classes, and both datasets have 500 training examples per class. Our experiment shows that on the CIFAR-100 (the smaller dataset), all the data-dependent regularization methods tend to perform well. When we switch to Tiny-ImageNet (the larger dataset), the (non-PatchUp) data-dependent regularization methods are still effective, but not necessarily close to the best method (PatchUp) as shown in Table~\ref{tab:tbl-cifar}  and~\ref{tab:tbl-tiny_imagenet}. 
{\renewcommand{\arraystretch}{1.}
\begin{table*}[hbt!]
\centering
    \begin{subtable}{1.0\linewidth}
        \resizebox{\linewidth}{!}{
\begin{tabular}{lllllllll}
\toprule
   & \multicolumn{2}{c}{\shortstack[l]{PreActResNet18}}                           &                      & \multicolumn{2}{c}{\shortstack[l]{PreActResNet34}}                           &                      & \multicolumn{2}{c}{\shortstack[l]{WideResnet-28-10}}                          \\ \cline{2-3} \cline{5-6} \cline{8-9}
   & \multicolumn{1}{c}{\shortstack[l]{Error}} & \multicolumn{1}{c}{\shortstack[l]{Loss}} & \multicolumn{1}{c}{} & \multicolumn{1}{c}{\shortstack[l]{Error}} & \multicolumn{1}{c}{\shortstack[l]{Loss}} & \multicolumn{1}{c}{} & \multicolumn{1}{c}{\shortstack[l]{Eror}} & \multicolumn{1}{c}{\shortstack[l]{Loss}} \\ \cline{2-3} \cline{5-6} \cline{8-9}
\shortstack[l]{No Mixup}                      &  $3.035 \pm 0.092$           & $0.138 \pm 0.004$ &      & $3.087 \pm 0.659$             & $0.164 \pm 0.008$ &                      & $2.833 \pm 0.081$            & $0.137 \pm 0.008$ \\ 
\shortstack[l]{Input Mixup ($\alpha$ = 1)}  &  $2.930 \pm 0.221$           & $0.233 \pm 0.016$ &      & $2.855 \pm 0.096$            & $0.223 \pm 0.024$ &                      & $2.643 \pm 0.161$             & $0.207 \pm 0.041$ \\ 
\shortstack[l]{ManifoldMixup ($\alpha$= 2)} &  {\ul{$2.436 \pm 0.056$}} & $0.157 \pm 0.062$ &      & {\ul{$2.423 \pm 0.428$}}  & $0.146 \pm 0.064$ &                      & $2.425 \pm 0.101$            & $0.157 \pm 0.029$ \\ 
\shortstack[l]{Cutout}                        &  $2.794 \pm 0.121$           & $0.122 \pm 0.010$ &      & $2.654 \pm 0.152$            & $0.114 \pm 0.007$ &                      & $2.475 \pm 0.148$            & $0.109 \pm 0.009$ \\ 
\shortstack[l]{DropBlock}                     &  $2.961 \pm 0.111$           & $0.134 \pm 0.005$ &      & $3.101 \pm 0.083$            & $0.158 \pm 0.006$ &                      & $2.732 \pm 0.055$             & $0.132 \pm 0.002$ \\ 
\shortstack[l]{CutMix}                        &  $3.040 \pm 0.054$          & $0.135 \pm 0.031$ &       & $2.658 \pm 0.049$            & $0.121 \pm 0.005$ &                      & $2.433 \pm 0.045$            & $0.110 \pm 0.003$ \\ 
\shortstack[l]{Puzzle Mix}                        &   $2.657 \pm 0.042$   & $0.113 \pm 0.003$ &         & $2.451 \pm 0.075$            & $0.111\pm 0.002$ &                      &   $2.432 \pm 0.068$          & $0.107 \pm 0.002$ \\
\shortstack[l]{\softpatchup{}}                &  $2.551 \pm 0.056$           & $0.129 \pm 0.023$ &      &  $2.467 \pm 0.081$               & $0.111 \pm 0.005$   &                      & {$\mathbf{2.081 \pm 0.066}$} & $0.111 \pm 0.010$ \\ 
\shortstack[l]{\hardpatchup{}}                &  {$\mathbf{2.286 \pm 0.084}$} & $0.107 \pm 0.004$ &     & {$\mathbf{2.123 \pm 0.024}$} & $0.101 \pm 0.007$ &                      & {\ul{$2.088 \pm 0.061$}}  & $0.105 \pm 0.012$ \\ \bottomrule
\end{tabular}}
\end{subtable}
 \captionsetup[subtable]{position = top}
    \captionsetup[table]{position=top}
    \caption{Error rates comparison on SVHN. We run experiments five times to report the mean and the standard deviation. The best performance result is shown in bold, second best is underlined. The lower number is better.}
    \label{tab:tbl-svhn}
\end{table*}}

{\renewcommand{\arraystretch}{1.1}
\begin{table}[htpb!]
\centering
    \begin{subtable}{1.0\linewidth}
        \resizebox{\linewidth}{!}{
        \begin{tabular}{lllllll}
        \toprule
        \shortstack[l]{PreActResNet18}                                   & \shortstack[l]{Test Error (\%)}   & \shortstack[l]{Test NLL}  &     & PreActResNet18                                   & \shortstack[l]{Test Error (\%)} & \shortstack[l]{Test NLL}\\ \cline{1-3} \cline{5-7}
        \shortstack[l]{No Mixup}                         &  $4.800 \pm 0.135$              &  $0.184 \pm 0.004$      &     & \shortstack[l]{No Mixup}                         &  $24.622 \pm 0.358$           &  $1.062 \pm 0.017$     \\
        \shortstack[l]{Input Mixup ($\alpha$ = 1)}     &  $3.628 \pm 0.201$              &  $0.192 \pm 0.012$      &     & \shortstack[l]{Input Mixup ($\alpha$ = 1)}     &  $22.326 \pm 0.323$           &  $1.011 \pm 0.012$     \\
        \shortstack[l]{ManifoldMixup ($\alpha$ = 1.5)}    &  $3.388 \pm 0.048$              &  $0.147 \pm 0.016$      &     & \shortstack[l]{ManifoldMixup ($\alpha$ = 1.5)}    &  $21.396 \pm 0.384$           &  $0.931 \pm 0.008$     \\
        \shortstack[l]{Cutout}                           &  $4.218 \pm 0.046$              &  $0.158 \pm 0.005$      &     & \shortstack[l]{Cutout}                           &  $23.386 \pm 0.185$           &  $1.004 \pm 0.004$     \\
        \shortstack[l]{DropBlock}                        &  $5.038 \pm 0.147$              &  $0.185 \pm 0.005$      &     & \shortstack[l]{DropBlock}                        &  $25.022 \pm 0.259$           &  $1.067 \pm 0.016$     \\
        \shortstack[l]{CutMix}                           &  $3.518 \pm 0.898$              &  $0.131 \pm 0.002$      &     & \shortstack[l]{CutMix}                           &  $22.184 \pm 0.176$           &  $0.949 \pm 0.012$     \\
        \shortstack[l]{Puzzle Mix}                       &   $3.155 \pm 0.110$             &  $0.119 \pm 0.004$      &     &              \shortstack[l]{Puzzle Mix}              & $20.649 \pm 0.214$           &  $0.857 \pm 0.013$    \\
        \shortstack[l]{\softpatchup{}}                   &  {\ul{$2.956 \pm 0.119$}}    &  $0.169 \pm 0.031$      &     & \shortstack[l]{\softpatchup{}}                   &  {\ul{$19.950 \pm 0.180$}} &  $0.833 \pm 0.005$     \\
        \shortstack[l]{\hardpatchup{}}                   &  $\mathbf{2.918 \pm 0.131}$       &  $0.146 \pm 0.018$      &     & \shortstack[l]{\hardpatchup{}}                   &  $\mathbf{19.120 \pm 0.172}$  &  $0.748 \pm 0.013$     \\ \cline{1-3} \cline{5-7}
        \shortstack[l]{PreActResNet34} & & & & \shortstack[l]{PreActResNet34} \\ \cline{1-3} \cline{5-7}
        \shortstack[l]{No Mixup}                        &  $4.640 \pm 0.099$               &  $0.204 \pm 0.004$      &     & \shortstack[l]{No Mixup}                        &  $23.342 \pm 0.269$            &  $1.103 \pm 0.006$     \\
        \shortstack[l]{Input Mixup ($\alpha$ = 1)}    &  $3.260 \pm 0.075$               &  $0.175 \pm 0.004$      &     & \shortstack[l]{Input Mixup ($\alpha$ = 1)}    &  $21.000 \pm 0.440$            &  $0.950 \pm 0.019$     \\
        \shortstack[l]{ManifoldMixup ($\alpha$ = 1.5)}  &  $2.926 \pm 0.062$               &  $0.124 \pm 0.004$      &     & \shortstack[l]{ManifoldMixup ($\alpha$ = 1.5)}  &  $18.724 \pm 0.305$            &  $0.810 \pm 0.008$     \\
        \shortstack[l]{Cutout}                          &  $3.690 \pm 0.141$               &  $0.150 \pm 0.012$      &     & \shortstack[l]{Cutout}                          &  $22.420 \pm 0.075$            &  $1.043 \pm 0.001$     \\
        \shortstack[l]{DropBlock}                       &  $4.950 \pm 0.188$               &  $0.221 \pm 0.010$      &     & \shortstack[l]{DropBlock}                       &  $23.744 \pm 0.125$            &  $1.113 \pm 0.007$     \\
        \shortstack[l]{CutMix}                          &  $3.332 \pm 0.071$               &  $0.142 \pm 0.004$      &     & \shortstack[l]{CutMix}                          &  $19.944 \pm 0.141$            &  $0.907 \pm 0.008$     \\
        \shortstack[l]{Puzzle Mix}                       & $2.996 \pm 0.069$               &  $0.125 \pm 0.006$      &     &        \shortstack[l]{Puzzle Mix}                    &  $19.974\pm 0.225$         &  $0.893\pm 0.022$    \\
        \shortstack[l]{\softpatchup{}}                  &  {\ul{$2.570 \pm 0.062$}}     &  $0.108 \pm 0.005$      &     & \shortstack[l]{\softpatchup{}}                  &  {\ul{$18.630 \pm 0.153$} } &  $0.816 \pm 0.016$     \\
        \shortstack[l]{\hardpatchup{}}                  &  {$\mathbf{2.534 \pm 0.048}$}    &  $0.108 \pm 0.005$      &     & \shortstack[l]{\hardpatchup{}}      &  {$\mathbf{17.692 \pm 0.125}$} &  $0.758 \pm 0.016$     \\ \cline{1-3} \cline{5-7}
        \shortstack[l]{WideResNet-28-10} & & & & \shortstack[l]{WideResNet-28-10} \\ \cline{1-3} \cline{5-7}
        \shortstack[l]{No Mixup}                        &  $4.244 \pm 0.142$               &  $0.162 \pm 0.011$      &     & \shortstack[l]{No Mixup}                        &  $22.442 \pm 0.226$            &  $1.065 \pm 0.010$     \\
        \shortstack[l]{Input Mixup ($\alpha$ = 1)}    &  $3.272 \pm 0.353$               &  $0.191 \pm 0.018$      &     & \shortstack[l]{Input Mixup ($\alpha$ = 1)}    &  $18.726 \pm 0.149$            &  $0.854 \pm 0.013$     \\
        \shortstack[l]{ManifoldMixup ($\alpha$ = 1.5)}  &  $3.252 \pm 0.183$               &  $0.155 \pm 0.034$      &     & \shortstack[l]{ManifoldMixup ($\alpha$ = 1.5)}  &  $18.352 \pm 0.378$            &  $0.833 \pm 0.023$     \\
        \shortstack[l]{Cutout}                          &  $3.134 \pm 0.119$               &  $0.122 \pm 0.005$      &     & \shortstack[l]{Cutout}                          &  $20.164 \pm 0.351$            &  $0.931 \pm 0.016$     \\
        \shortstack[l]{DropBlock}                       &  $4.182 \pm 0.069$               &  $0.157 \pm 0.003$      &     & \shortstack[l]{DropBlock}                       &  $22.364 \pm 0.149$            &  $1.049 \pm 0.013$     \\
        \shortstack[l]{CutMix}                          &  $3.148 \pm 0.118$               &  $0.126 \pm 0.004$      &     & \shortstack[l]{CutMix}                          &  $18.316 \pm 0.185$            &  $0.839 \pm 0.020$     \\
        \shortstack[l]{Puzzle Mix}                       &   {\ul{$2.562 \pm 0.074$}}             &   $0.098
 \pm 0.002$     &     &       \shortstack[l]{Puzzle Mix}                     &       $17.528 \pm 0.224$     &   $0.757 \pm 0.006$   \\
        \shortstack[l]{\softpatchup{}}                  &  $2.606 \pm 0.052$   &  $0.132 \pm 0.029$      &     & \shortstack[l]{\softpatchup{}}                  &  {\ul{$16.726 \pm 0.110$}}  &  $0.722 \pm 0.017$     \\
        \shortstack[l]{\hardpatchup{}}                  &  {$\mathbf{2.528 \pm 0.065}$}    &  $0.114 \pm 0.014$      &     & \shortstack[l]{\hardpatchup{}}                  &  {$\mathbf{16.134 \pm 0.197}$} &  $0.660 \pm 0.017$     \\ \bottomrule
        \multicolumn{3}{c}{\shortstack[l]{Comparison on CIFAR-10}}                & \multicolumn{1}{c}{} & \multicolumn{3}{c}{\shortstack[l]{Comparison on CIFAR-100}}
        \end{tabular}}
\end{subtable}
 \captionsetup[subtable]{position = top}
    \captionsetup[table]{position=top}
    \caption{Image classification task error rates on CIFAR-10 and CIFAR-100. We run experiments five times to report the mean and the standard deviation. The best performance result is shown in bold, second best is underlined. The lower number is better.}
    \label{tab:tbl-cifar}
\end{table}}


\end{document}